\documentclass[journal]{IEEEtai}

\usepackage{times}
\usepackage{helvet}
\usepackage{courier}
\usepackage{graphicx}
\usepackage{caption}
\usepackage{newfloat}
\usepackage{listings}

\usepackage[utf8]{inputenc} 
\usepackage[T1]{fontenc}    
\usepackage{url}            
\usepackage{booktabs}       
\usepackage{nicefrac}       
\usepackage{microtype}      
\usepackage{xcolor}         
\usepackage{multirow, booktabs}
\usepackage{float}
\usepackage{subcaption}
\usepackage{diagbox}

\usepackage{amsmath,amsfonts}
\usepackage{algorithmic}
\usepackage{algorithm}

\usepackage{array}
\usepackage{textcomp}
\usepackage{stfloats}
\usepackage{verbatim}
\usepackage{graphicx}
\usepackage{cite}
\usepackage{hyperref}
\begin{document}

\title{Query Attack by Multi-Identity Surrogates}

\author{Sizhe Chen, Zhehao Huang, Qinghua Tao, Xiaolin Huang,~\IEEEmembership{Senior~Member,~IEEE}
\thanks{This work was partially supported by National Natural Science Foundation of China (61977046), Shanghai Science and Technology Program (22511105600), and Shanghai Municipal Science and Technology Major Project (2021SHZDZX0102).} \protect\\
\thanks{S. Chen, Z. Huang, and X. Huang are with the Department of Automation, and the Institute of Medical Robotics, Shanghai Jiao Tong University, and also with the MOE Key Laboratory of System Control and Information Processing, 800 Dongchuan Road, Shanghai, 200240, P.R. China. e-mails: sizhe.chen@berkeley.edu, \{kinght\_h, xiaolinhuang\}@sjtu.edu.cn. Q. Tao is with ESAT-STADIUS, KU Leuven, Belgium. email: qinghua.tao@esat.kuleuven.be. Corresponding author: Xiaolin Huang.}
\thanks{Manuscript received November 21, 2022, revised February 21, 2023, accepted March 12, 2023.}}

\markboth{IEEE TRANSACTIONS ON ARTIFICIAL INTELLIGENCE}%
{Chen \MakeLowercase{\textit{et al.}}: Query Attack by Multi-Identity Surrogates}

\IEEEpubid{}

\maketitle

\begin{abstract}
  Deep Neural Networks (DNNs) are acknowledged as vulnerable to adversarial attacks, while the existing black-box attacks require extensive queries on the victim DNN to achieve high success rates. For query-efficiency, surrogate models of the victim are used to generate transferable Adversarial Examples (AEs) because of their \emph{Gradient Similarity (GS)}, i.e., surrogates' attack gradients are similar to the victim's ones. However, it is generally neglected to exploit their similarity on outputs, namely the \emph{Prediction Similarity (PS)}, to filter out inefficient queries by surrogates without querying the victim. To jointly utilize and also optimize surrogates' GS and PS, we develop QueryNet, a unified attack framework that can significantly reduce queries. QueryNet creatively attacks by multi-identity surrogates, i.e., crafts several AEs for one sample by different surrogates, and also uses surrogates to decide on the most promising AE for the query. After that, the victim's query feedback is accumulated to optimize not only surrogates' parameters but also their architectures, enhancing both the GS and the PS. Although QueryNet has no access to pre-trained surrogates' prior, it reduces queries by averagely about an order of magnitude compared to alternatives within an acceptable time, according to our comprehensive experiments: 11 victims (including two commercial models) on MNIST/CIFAR10/ImageNet, allowing only 8-bit image queries, and no access to the victim's training data. The code is available at \url{https://github.com/Sizhe-Chen/QueryNet}.
\end{abstract}

\begin{IEEEImpStatement}
Deep Neural Networks (DNNs) have been validated as vulnerable, while they are much harder to attack in real-world black-box settings. Existing methods require hundreds of queries on the victim DNN to perform an effective attack. By proposing a new perspective that gradient and prediction similarity are fundamentally different, this paper fosters the query-based attackers via stealing the victim model using neural architecture search with a novel sample evaluation. Our unified attack framework, namely QueryNet, reduces the average query times to break a DNN by an order of magnitude from current baselines. Because practical DNNs are inevitably subject to intense queries, our work reveals the significant adversarial threat in DNN applications and advocates for attention to securing DNNs in deployment. Our view on DNN similarities may also inspire future work on attacks and defenses by surrogate models.
\end{IEEEImpStatement}


\section{Introduction}
\IEEEPARstart{A}{s} Deep Neural Networks (DNNs) develop into the mainstream tools in pattern recognition for their high performance, their vulnerabilities become non-negligible \cite{8844865, 9206141, 9573256, 9451562, 9470919}. A widely-known fact is that DNNs are sensitive to maliciously crafted imperceptible perturbations \cite{goodfellow2014explaining, moosavi2016deepfool, carlini2017towards, madry2017towards}, and the resulted \emph{Adversarial Examples (AEs)} could fool well-trained \emph{victim} DNNs \cite{tang2021adversarial, 9018372, 9632406, 9325048,9472988,9524508}. In real-world scenarios, the attackers generally have no access to the architecture, parameters, training data, and algorithmic details of the victim, and thus the so-called \emph{black-box attacks} \cite{papernot2017practical, chen2017zoo, ilyas2018prior} are needed, which commonly depend on querying for victim's outputs to iteratively modify the sample. Existing black-box query attacks already have very high success rates, so currently, the main efforts are devoted to reducing the number of queries \cite{cheng2019improving, al2019sign, guo2019simple, andriushchenko2020square}, which is generally based on the similarity between the victim and the \emph{surrogate} models at attackers' hands. Due to the similarity, it is expected that white-box AEs on surrogates can \emph{transfer} to hurt the victim as well \cite{goodfellow2014explaining, dong2018boosting, xie2019improving}.

Successful transfer attacks from the surrogates rely on the \emph{Gradient Similarity (GS)} between surrogates and the victim \cite{guo2019subspace, du2019query, yang2020learning}, because an AE is produced by adding surrogate's gradients to the original sample. For example, the attack on attention \cite{chen2020universal, chen2022relevance}, a method with high transferability, essentially relies on the similarity in the attention heap map, calculated based on gradients. However, GS is only one aspect and there is another critical but mostly neglected factor in attacks, which is the \emph{Prediction Similarity (PS)}. Specifically, if the surrogates' predictions are similar to the victim's ones, they could filter out inefficient AEs, and thus prevent queries of less promising samples \cite{ma2020metasimulator}. Notice that GS and PS are not naturally consistent, e.g., \cite{zhou2020dast} obtains a surrogate with high GS by querying and training, and its AEs transfer to the victim well, but the surrogates' accuracy is very low, indicating that the PS is not good. Generally speaking, existing methods have not taken the full use of the surrogates by jointly exploiting their GS and PS, resulting in unsatisfactory attack performance, e.g., over 500 average queries in attacking ImageNet \cite{deng2009imagenet} models under $\ell_2=5$.

\begin{figure*}
	\centering
	\includegraphics[width=\hsize]{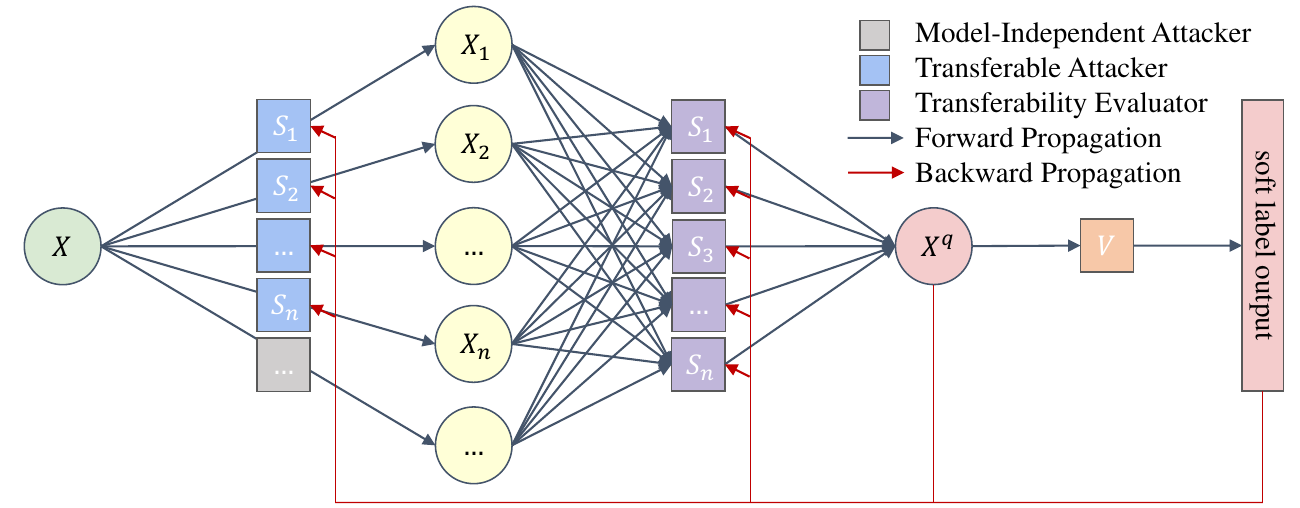}
	\caption{QueryNet forward propagates to obtain query samples $X^q$ towards the victim $V$ (dark line), and the query pairs back propagate to optimize the surrogates (red line). In the forward propagation, we develop a novel manner to use multi-identity surrogates, i.e., not only as transferable attackers (blue) to craft candidate AEs $X_i$ along with other attackers (grey), but also as transferability evaluators (purple) to decide on the most promising candidate AE $X^q$ for the query. In the backward propagation, QueryNet creatively adopts NAS to train surrogates, increasing their GS and PS with $X^q$ labeled by victim's outputs. By the unified exploitation and improvement of both surrogate's GS and PS, QueryNet reduces queries by about an order of magnitude.}
	\label{fig:intro}
\end{figure*}

In this paper, we propose QueryNet to take advantage of surrogates' multiple identities so that their GS and PS are jointly exploited and improved. QueryNet is a unified framework for score-based query attacks, and its overall structure is illustrated in Fig. \ref{fig:intro}. From an original sample, QueryNet crafts several transferable \emph{candidate} AEs by surrogates (transferable attackers) and also uses surrogates (transferability evaluators) to decide on the most promising AE for the query. By novel usage of surrogates with multiple identities, QueryNet generates effective queries from diverse candidates, reducing extensive query times. Furthermore, QueryNet also increases query-efficiency by creatively adopting neural architecture search (NAS) to optimize surrogates. Reasons herein are that QueryNet does not take pre-trained surrogate's knowledge, and attaining both high PS and GS in this case is challenging, requiring precise approximation to the victim from both the parameter and architecture aspect.


We comprehensively evaluate QueryNet with various state-of-the-art (SOTA) attacks proposed after 2019 under real-world settings: no access to the training data and only 8-bit image queries is allowed. Our experiments include 11 victims, two of which are commercial black-box models from the Baidu EasyDL platform \cite{easydl}. Overall, QueryNet requires very few queries, reducing queries by 45\% to 97\% from the excellent SOTA Square attack \cite{andriushchenko2020square} baseline. For example, when attacking WideResNet \cite{ZagoruykoK16} under $\ell_2=3$, QueryNet reduces the average queries from 192.0 (baseline) to 5.3 on MNIST \cite{lecun2010mnist} and from 70.0 to 9.5 on CIFAR10 \cite{krizhevsky2009learning}. Besides, QueryNet uses only 24.0 average queries for definite fooling under $\ell_\infty=0.05$ on MNASNet \cite{tan2019mnasnet}, an ImageNet \cite{deng2009imagenet} model. QueryNet achieves amazing query-efficiency within an acceptable amount of time due to our optimization and paralleling in implementation. Even with NAS, QueryNet only averagely consumes 0.15 to 0.65 seconds to craft an AE for MNIST and CIFAR10 victims under the common $\ell_\infty$ attack bounds.

We summarize our contributions as follows.
\begin{itemize}
    \item We propose QueryNet, a unified attack framework, where surrogates' GS and PS are jointly exploited in forward propagation and optimized in backward propagation.
    \item We design a novel AE evaluation mechanism in QueryNet's forward propagation and creatively optimize surrogates by NAS in the backward propagation, forming an extremely query-efficient score-based attack.
    \item We comprehensively evaluate QueryNet under real-world settings, demonstrating that QueryNet reduces queries by averagely about an order of magnitude compared to recent SOTA methods while maintaining the highest fooling rate.
\end{itemize}

The rest of this paper is organized as follows. In Section \ref{related}, we will briefly introduce adversarial attacks, especially black-box query attacks and the surrogate models. The QueryNet is described in detail in Section \ref{method}. Section \ref{experiment} evaluates QueryNet with baselines, presenting visual and numerical results to show QueryNet's superiority. In Section \ref{conclusion}, a conclusion is given to end this paper with potential future work.

\section{Related Work}\label{related}
Adversarial attacks craft human-imperceptible perturbations on the input to mislead the victim DNN to make incorrect predictions. Since its proposal \cite{szegedy2013intriguing}, many white-box attacks are developed \cite{goodfellow2014explaining, moosavi2016deepfool, carlini2017towards, madry2017towards}, using the gradients of the victim to adversarially update the input, i.e., to increase the training loss, so that the victim produces false outputs with high confidence. White-box attacks generally achieve high success rates since the victim's complete information is accessible, otherwise one has to adopt black-box attacks, which are generally categorized into transfer attacks \cite{dong2018boosting, xie2019improving, dong2019evading, guo2020backpropagating, chen2020universal} and query attacks \cite{papernot2017practical, chen2017zoo, ilyas2018prior, cheng2019improving, al2019sign, guo2019simple, andriushchenko2020square}. Transfer attacks conduct white-box attacks on the surrogate model of the victim and expect the yielded AEs to hurt the victim, which is referred to as the attack transferability. The surrogate and the victim are both accurate, and thus are very similar in the prediction, but the attack transferability remains relatively low. In contrast, query attacks target a specific victim and update AEs iteratively according to the victim's output feedback. They generally enjoy high success rates, but the number of queries leaves much to be reduced. According to the accessibility of the victim's output, query attacks are further divided into score-based attacks \cite{papernot2017practical, chen2017zoo, ilyas2018prior, cheng2019improving, al2019sign, guo2019simple, andriushchenko2020square} and decision-based ones \cite{brendel2018decision, chen2020hopskipjumpattack, cheng2019sign}. The former ones provide the probability vectors of the victim's output for attackers, while the latter ones only allow access to the victim's decisions, i.e., the top-1 predictions.

Score-based query attacks are our focus. Early studies try to estimate the victim's gradients by additional queries around the sample \cite{chen2017zoo, bhagoji2018practical} inspired by white-box attacks, but this strategy is very query-intensive. Besides gradient estimation, fast pure random-search methods are also validated effective \cite{guo2019simple, andriushchenko2020square, li2020projection}, which greedily update AEs only if the perturbation makes the sample more adversarial to the victim. Parallel methods introduce the surrogate model in transfer attacks to generate the transfer-prior, i.e., the initial adversarial perturbations \cite{cheng2019improving}, before querying the victim. Based on this idea, \cite{guo2019subspace} regards surrogates' transfer-prior as subspaces to reduce the query searching space; \cite{du2019query} adopts meta-learning to approximate the victim; \cite{yang2020learning} proposes to learn the victim's (estimated) gradients via high-order computation graph. The aforementioned methods only exploit surrogates' GS towards the victim to generate AEs. Recently, \cite{ma2020metasimulator} uses surrogates' PS for virtual queries on the victim, but it does not consider the GS and is resource-consuming due to the meta-training. Moreover, transfer attacks and random-search query attacks are still developing separately to the best of our knowledge.

Surrogates in query attacks can be trained differently. Most works train surrogates in a traditional way, i.e., decreasing the cross-entropy loss on the victim's training data \cite{cheng2019improving, ma2020switching} or external data \cite{guo2019subspace}. Recently, \cite{du2019query, ma2020metasimulator} propose to adopt meta-training to imitate the victim; \cite{yang2020learning} alters the loss to explicitly learn the victim's (estimated) gradients; \cite{zhou2020dast} trains a high-transferable AE generator from queries of no natural data; \cite{feng2020boosting} models the conditional adversarial distribution by a conditional generative flow. The methods above consider the query data pair besides the victim's training data. Comparatively, we only resort to query pairs. Also, we introduce NAS in training, because NAS optimizes the network in a larger space, i.e., modifying architectures besides parameters, which matches our need to precisely approximate the victim. Early NAS methods \cite{zoph2017neural, zoph2018learning, real2019regularized} consume much GPU resources. Differential ARchitecture Search (DARTS \cite{liu2019darts}) greatly reduces the cost of NAS by relaxing the search space to be continuous. Partially-Connected DARTS (PC-DARTS \cite{xu2020pc}) samples a small component of super-network to reduce the memory consumption without hurting the performance.

\section{The proposed QueryNet}\label{method}
In this section, we first introduce our notations and terminologies, then state our motivation to consider GS and PS by analyzing existing methods. After that, we present the design of QueryNet, especially the novel AE mechanism in the forward path and the creative NAS optimizing in the backward path. Finally, we summarize our QueryNet and the iterative query strategy in Alg. \ref{amida} and Alg. \ref{procedure} respectively.
\subsection{Preliminaries}
The black-box attack problem is formulated as below. For an original $\boldsymbol x^\mathrm{org}$ that is correctly predicted by the victim as its ground truth class $y^{\mathrm{org}}$ by a $K$-classifier victim $V$, the goal of the untargeted score-based query attack is to achieve
\begin{eqnarray}\label{attack}
\begin{split}
\left\{
\begin{array}{ll}
 {{\arg \max} ~(\mathcal{V}(\boldsymbol x^\mathrm{adv})) \neq y^{\mathrm{org}}},  \\ 
{\|\boldsymbol x^\mathrm{adv} - \boldsymbol x^\mathrm{org}\|_p \leq \varepsilon}, \\
{\# \mathcal{V}(\boldsymbol x)\leq N,}
\end{array}
\right.
\end{split}
\end{eqnarray}
where $\mathcal{V}(\boldsymbol x): \mathbb R^d \mapsto \mathbb R^K$ represents the victim DNN with the prediction $[\mathcal{V}_1(\boldsymbol x), \ldots, \mathcal{V}_K(\boldsymbol x)]^T$.

The first condition in (\ref{attack}) means fooling the victim. In our QueryNet, we use the following margin loss
\begin{eqnarray}\label{loss}
\mathcal{L}(\mathcal{V}(\boldsymbol x), y)=\mathcal{V}_{y}(\boldsymbol x)-\max _{k \neq y} \mathcal{V}_{k}(\boldsymbol x),
\end{eqnarray}
which stands for the smallest difference between victim's output on the correct class and that on incorrect classes, and this is a simplified C\&W attack loss \cite{carlini2017towards}. If $\mathcal{L}(\mathcal{V}(\boldsymbol x), y^{\mathrm{org}}) < 0$, the sample $\boldsymbol x$ becomes an AE, and the attacker could also minimize surrogate's margin loss to craft transferable examples. Since the query times for different samples vary a lot \cite{ozbulak2021selection}, the attack performance is generally evaluated on several images. Thus, in descriptions below, we uniformly denote a set of AEs as $X^\mathrm{adv}=\{\boldsymbol x^{\mathrm{adv}}_k\}$, and its counterpart original samples as $X^\mathrm{org}$. Similarly, $\mathcal{V}(X)$ contains all predictions, and $L=\mathcal{L}(\mathcal{V}(X), Y)$ is the loss.

The second condition in (\ref{attack}) guarantees the attack imperceptibility, i.e., the perturbations are $\ell_p$-norm bounded by $\varepsilon$, where $p$ is commonly set as $2$ or $\infty$. The last condition in (\ref{attack}) is the budget for queries, meaning that the victim's output $\mathcal{V}(\boldsymbol x)$ is allowed to be accessed for no more than $N$ times.

Besides our notations, we also summarize terminologies appeared in the introduction for a clear subsequent presentation.
\begin{itemize}
    \item gradient similarity (GS): A surrogate has high GS if its gradients are similar to the victim, which is quantified by the attack transferability of its AEs towards the victim.
    \item transferable attacker: A surrogate is a transferable attacker if we use its gradients to generate AEs for the victim, i.e., exploiting its GS towards the victim.
    \item prediction similarity (PS): A surrogate has high PS if its predictions are similar to the victim, which is quantified by its prediction consistency with the victim.
    \item transferability evaluator: A surrogate is a transferability evaluator if we use it to evaluate and estimate the adversarial transferability of AEs, i.e., exploiting its PS towards the victim.
    \item multiple identities: refer to that surrogates are multiply exploited as transferable attackers and transferability evaluators in QueryNet.
    \item candidate AEs: crafted from the same original samples by different attackers, sent to the transferability evaluators as candidates for query samples.
\end{itemize}

\subsection{Motivation}
To attack in a query-efficient manner, surrogate models of the victim DNN have been adopted. The basic idea herein is to imitate the victim by the complete accessible surrogates at the attacker's hands. The imitation is based on the similarity between the surrogate and the victim, which could be roughly divided into three levels.

\textbf{Complete PS:} The surrogate predicts exactly the same as the victim, leading to complete PS and GS. This unobtainable setting degrades black-box attacks to white-box ones.

\textbf{High PS, Low GS:} The surrogates' outputs are quite similar to that of the victim, but their gradients differ distinctively. A direct example is the transfer attack, where surrogates share similar predictions with the victim due to the training on the same data, but surrogate's AEs may not transfer to the victim since their GS is low. Although in this case, the attack transferability is not good, the high PS makes it promising to use the surrogate to decide whether an AE would hurt the victim or not \cite{ma2020metasimulator}. This setting is achievable by training surrogates with victim's training data or sufficient data labeled by the victim, which is similar to knowledge distillation \cite{hinton2015distilling} or model stealing attacks \cite{tramer2016stealing, orekondy2019knockoff}.

\textbf{Low PS, High GS:} The attacker has little knowledge about the victim's exact predictions, but could accurately guess its gradients, which could be realized without \cite{chen2017zoo, bhagoji2018practical} or with \cite{cheng2019improving, guo2019subspace} surrogates. Under this setting, white-box attacks on the surrogates are promising to craft transferable and efficient AEs towards the victim. But since the PS is low, the exact outputs of the victim cannot be well known and numerous queries are still needed.

According to the description above, we roughly categorize existing attacks in Table \ref{methodcompare}. Commonly, the attack methods use one surrogate. More importantly, every surrogate only carries a single identity, i.e., is exploited either as an attacker due to their GS (column 3), or as a virtual predictor for their PS (column 4). Accordingly, the surrogate is utilized from a limited perspective, leading to great room for improving the attack performance. In this regard, it is imperative to develop a method that exploits multi-identity surrogates for both their GS and PS, and improves their GS and PS by accurate self-adjustments.

\begin{table}
\caption{Comparison with existing methods}
\label{methodcompare}
\centering
\begin{tabular}{c|c|c|c}
\toprule
\multirow{2}*{Method} & trainable & attack by $S$ & predict by $S$  \\
&  components  & (exploit GS) & (exploit PS)  \\  \midrule
Bandits \cite{ilyas2018prior} & $\times$  & $\times$ & $\times$ \\
NES \cite{ilyas2018black} & $\times$  & $\times$ & $\times$ \\
RGF \cite{nesterov2017random} & $\times$  & $\times$ & $\times$ \\
\midrule
Transfer \cite{dong2018boosting, xie2019improving}  & $\times$   & single & $\times$ \\
P-RGF \cite{cheng2019improving} & $\times$ & single & $\times$ \\
\midrule
DaST \cite{zhou2020dast} & $S$ & single & $\times$ \\
Meta \cite{du2019query} & $S$   & single & $\times$ \\
LeBA \cite{yang2020learning}  & $S$   & single & $\times$ \\
\midrule
DPD \cite{zhang2020dual}  & distribution & single & $\times$ \\
CAD \cite{feng2020boosting} & distribution & multiple & $\times$ \\
Subspace \cite{guo2019subspace}& $\times$ &  multiple & $\times$ \\
Simulator \cite{ma2020metasimulator} & $S$  & $\times$ & \checkmark \\
\midrule
QueryNet (ours) &  $S, \boldsymbol w$ &  multiple & \checkmark \\
\bottomrule
\end{tabular}
\end{table}

\subsection{Craft queries by QueryNet: exploit GS and PS}
In this paper, we propose a novel attack framework QueryNet. As illustrated in Fig. \ref{fig:intro}, QueryNet crafts the query samples by multi-identity surrogates in the forward propagation, and uses NAS to adjust surrogates by the victim's feedback in the backward propagation.

Concretely, QueryNet first exploits surrogates' GS and PS by generating various candidate AEs for one sample, and then selecting one of them with the highest (estimated) attack transferability $X^q$ for the query, that is, the sample with the lowest surrogate's margin loss. This novel usage of surrogates yields diverse AEs but only queries the victim with the most effective ones, reducing extensive query times. After that, QueryNet improves surrogates' GS and PS by creatively adopting NAS to optimize the surrogates $S$, so that we could attack with better surrogates in the next query.

In the forward propagation, QueryNet first crafts diverse candidate AEs $X_i$ by attacker $\mathcal{A}_i$ as $X_i = \mathcal{A}_i^\varepsilon(X, \cdot)$. 
After that comes the most significant and distinctive part of QueryNet, the self-evaluation of potential candidate AEs. The idea is to estimate an AE's transferability towards the victim by its transferability towards all surrogates, which is accessible and quantifiable as a weighted sum of surrogates' loss values. Accordingly, the overall forward propagation to craft the $k^\text{th}$ query sample at one iteration could be expressed as
\begin{eqnarray}\label{update}
\begin{split}
X^q_k & = X_{a_k,k} = \mathcal{A}_{a_k}^{\varepsilon}(X_k, \cdot),\\
a_k &=\underset{i}{\operatorname{argmin}} \sum\limits_{j=1}^n w_j \mathcal{L}(S_j(\mathcal{A}_i^{\varepsilon}(X_k, \cdot)), Y_k^\mathrm{org}).
\end{split}
\end{eqnarray}
The first equation means that the query sample $X^q_k$ is from attacker $a_k$, because, according to the second equation, it produces the AE with the lowest weighted surrogates' loss, i.e., the highest transferability toward surrogates. Here $i, j$ stand for the indexes of the attacker, evaluator respectively. $w_j$ is the evaluation weight for surrogate $S_j$, and $\boldsymbol a$ is a vector containing indexes of the samples.

In this way, QueryNet constructs a novel strategy to exploit not only surrogates' GS to craft AEs but also their PS to filter out ineffective queries. Next, we discuss how we creatively improve GS and PS by query feedback to further reduce queries.

\subsection{Self-adjustments in QueryNet: improve GS and PS}
In the backward propagation, QueryNet adjusts its surrogates by the victim's output. Instead of using pre-trained surrogates prior knowledge \cite{cheng2019improving, guo2019subspace, yang2020learning}, our surrogates approximate $V$ by training only on past query pairs $D=(X^q, \mathcal{V}(X^q))$ to minimize an MSE loss as in \cite{du2019query, yang2020learning, ma2020metasimulator}, i.e.,
\begin{eqnarray}\label{midaproblem}
\begin{split}
\min\limits_\mathcal{S} ~\sum\limits_{j=1}^n \|\mathcal{S}_j(\boldsymbol x) - \mathcal{V}(\boldsymbol x)\|_2^2~, ~~ \boldsymbol x \in D. \\
\end{split}
\end{eqnarray}
Unlike existing methods \cite{du2019query, yang2020learning}, QueryNet exploits multi-identity surrogates by using them as transferable attackers and transferability evaluators, pursing high similarity on both prediction and gradient, which raises a higher requirement on surrogate’s approximation of the victim than focusing on one part. Therefore, in the self-adjustment of QueryNet, we alter also surrogates' architectures by NAS. NAS optimizes DNN's parameters and architecture by the training data and validation data respectively. The query pairs are generated naturally as the attack proceeds. For example, in attacking 9500 correctly-classified samples, we first add perturbations to them and query a victim. Then according to the query feedback, 500 adversarial examples are successfully crafted, so we repeat the same query process for the rest of 9000 samples, and get 9500+9000=18500 query pairs in total after 2 attack iterations. In the later iterations, more query pairs would be similarly generated and recorded for optimizing QueryNet surrogates' parameters and architecture by randomly sampling from all query pairs. This step is denoted as $S' = \mathcal{O}(S, D)$, meaning that the surrogate is trained from $S$ to $S'$ until convergence. In this paper, PC-DARTS \cite{xu2020pc} is adopted here for its high performance and efficiency on time and memory.

\begin{figure}
    \centering
    \includegraphics[width=\hsize]{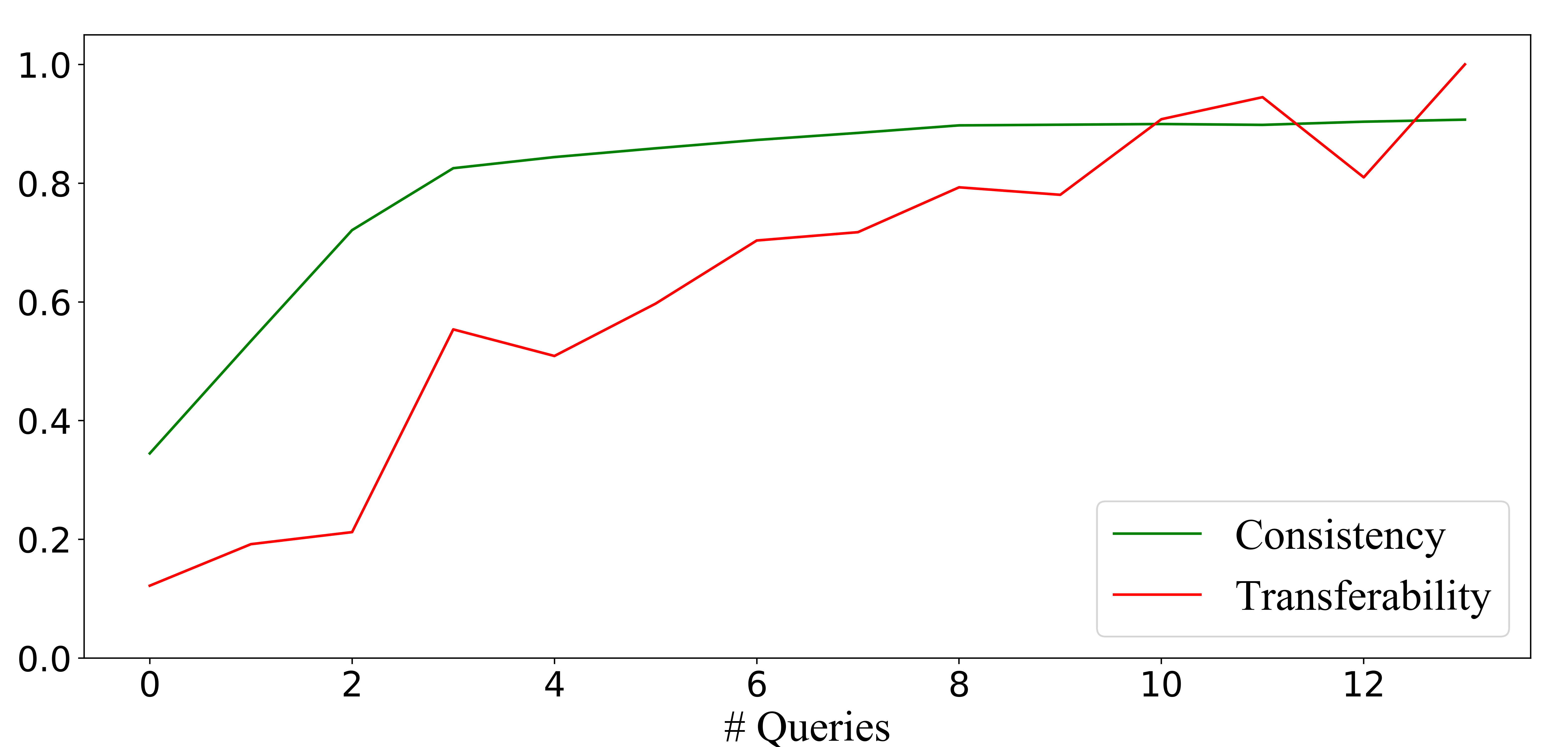}
    \caption{The trend of surrogates' consistency (PS), the attack transferability (GS) after every attack iteration (querying all 10K test samples). The consistency refers to the rate of similar predictions made by $V$ and $S$ on all past queries. The transferability is calculated by the victim's error rates on the surrogate's AEs from all original test samples, and we normalize them to make the largest value as 1 for a clear demonstration. The results are generated from attacking DenseNet40 \cite{huang2017densely} on MNIST with $\ell_2=3$.}
    \label{fig:psgs}
\end{figure}

Through our novel NAS training for attackers, the PS and GS of surrogates both increase during the attack, leading to a precise approximation for the victim. We demonstrate this by an example in Fig. \ref{fig:psgs}. The PS (indicated by consistency) and GS (indicated by transferability) quickly increase after a few attack iterations, validating the effectiveness of our NAS optimization. The query time here means average query for each sample, and since attackers generally produce many AEs \cite{ma2020metasimulator, yang2020learning, zhang2020dual}, the data is sufficient for convergence on a good solution.

\begin{algorithm}
  \caption{QueryNet} 
  \label{amida}
  \begin{algorithmic}[1]
    \REQUIRE{Pre-update samples $X$, labels $Y$, attack bound $\varepsilon$, victim $V$, surrogates $S$, past query pairs $D$, evaluation weights $\boldsymbol w$, NAS method $\mathcal{O}$, attackers $\mathcal{A}$.}
    \ENSURE{query samples $X^q$}
    \STATE $X_i = \mathcal{A}_i^\varepsilon(X, \cdot)$~~ \emph{\# craft candidate AEs with or without $S$}
    \STATE $a_k =\underset{i}{\operatorname{argmin}} \sum_{j=1}^n w_j \mathcal{L}(S_j(\mathcal{A}_i^{\varepsilon}(X_k, \cdot)), Y_k^\mathrm{org})$ 
    \STATE $X^q_k = X_{a_k, k}$~ \emph{\# select AEs with the highest transferability}
    \STATE $D = D \cup (X^q, \mathcal{V}(X^q))$~~\emph{\# store the query feedback} 

        \STATE $S_i = \mathcal{O}(S_i, D) ~\textbf{for}~ i = 1 \to n$ ~~\emph{\# NAS on $S$ by (\ref{midaproblem})} 
        \STATE Update $\boldsymbol w$ according to (\ref{weights}) \emph{\# estimate surrogates' quality}
    \RETURN $X^q$ 
    \end{algorithmic}
\end{algorithm}

\subsection{QueryNet algorithm: a summary}
The aforementioned QueryNet, including the forward propagation and backward propagation, is summarized in Alg. \ref{amida}. In the forward propagation, we design a novel strategy to evaluate diverse candidate AEs, which are generated by various attackers. Among them, surrogate attackers are used for their GS to attack by FGSM \cite{goodfellow2014explaining}. Besides, two model-independent attacks are also involved to diversify the attack, which are the excellent Square attack \cite{andriushchenko2020square} and Square+, a variant of Square attack designed by us. Details of all attackers are presented in Appendix \ref{attackdetails}. In the backward propagation, we creatively adopt NAS to optimize surrogates by the query pairs. In this way, surrogates' GS and PS increase quickly during the attack as in Fig. \ref{fig:psgs}. Besides improving the quality of a surrogate, we also pay attention to calibrating its quality. Thus, we also learn surrogates' evaluation weights $\boldsymbol w$ by the victim's feedback, which is specified in Appendix \ref{midaalg}.


Before we have discussed how to attack in one iteration by QueryNet, next we present the iterative query attack strategy for QueryNet in Alg. \ref{procedure}. The procedure starts by training randomly initialized surrogates with query pairs of original samples $(X^\mathrm{org}, \mathcal{V}(X^\mathrm{org}))$. At each attack iteration, we only focus on unsuccessful AEs $X$, i.e., those correctly predicted by $V$ (line 6 in Alg. \ref{procedure}). Then we use QueryNet to craft $X^q$ to query the victim. Finally, we update our AEs by those that become more adversarial to the victim (line 8 in Alg. \ref{procedure}), i.e., always hold the ``most adversarial'' examples with the smallest victim's loss.

\begin{algorithm}[t]
  \caption{Iterative Query Strategy for QueryNet}
  \label{procedure}
  \begin{algorithmic}[1]
    \REQUIRE{Original samples $X^{\mathrm{org}}$, labels $Y^{\mathrm{org}}$, attack bound $\varepsilon$, maximum query $N$, victim $V$, number of surrogates $n$, evaluation weights $\boldsymbol w$, NAS method $\mathcal{O}$, attackers $\mathcal{A}$.}
    \ENSURE{adversarial examples $X^{\mathrm{adv}}$}

    \STATE $X^{\mathrm{adv}} \gets X^{\mathrm{org}}, ~~L^{\mathrm{adv}} \gets \mathcal{L}(\mathcal{V}(X^{\mathrm{adv}}),Y^{\mathrm{org}})$
    \STATE $D \gets (X^{\mathrm{adv}}, \mathcal{V}(X^{\mathrm{adv}}))$~~ \emph{\# store all past query pairs}
    \STATE $S = [S_1, S_2,...S_n] \gets random~initialization$
    \STATE $S_i = \mathcal{O}(S_i, D) ~\textbf{for}~ i = 1 \to n$ ~~ \emph{\# train by initial queries}

    \REPEAT
        \STATE $F= \{k: L_k^{\mathrm{adv}} > 0\}, (X,Y,L) = (X^{\mathrm{adv}}_{F}, Y^{\mathrm{org}}_{F}, L^{\mathrm{adv}}_{F})$
        \STATE Get query samples $X^q$ by QueryNet (Alg. \ref{amida})
        \STATE $I = \{k: L^q_k < L_k \}, ~(X^{\mathrm{adv}}_{F_I}, ~L^{\mathrm{adv}}_{F_I}) = (X^q_I, ~L^q_I)$ 
    \UNTIL{$N$ times or $X^{\mathrm{adv}}_{F}$ is $\emptyset$}

    \RETURN $X^{\mathrm{adv}}$
    \end{algorithmic}
\end{algorithm}

By the iterative implementation, QueryNet's novel AE evaluation mechanism and creative NAS training dramatically increase the query-efficiency by exploiting and improving multi-identity surrogates' GS and PS in a unified framework, which is validated by our extensive experiments below.

\section{Experiments}\label{experiment}
\subsection{Setup}\label{sec:setup}
\textbf{Baselines.} We evaluate QueryNet along with several SOTA attacks proposed after 2019, which are Bandits \cite{ilyas2018prior}, Subspace \cite{guo2019subspace}, LeBA \cite{yang2020learning}, PPBA \cite{li2020projection}, SimBA \cite{guo2019simple}, and Square \cite{andriushchenko2020square}. Notice that most of the attacks are originally tested on CIFAR10 and we select the most efficient ones, including Bandits, PPBA, Subspace, and Square for MNIST and ImageNet. For the compared methods, we generally inherit the default hyper-parameters from their papers, seeing Appendix \ref{hyper}. Results show that QueryNet, using only randomly-initialized surrogates, outperforms all baselines, even those using pre-trained surrogates' prior knowledge \cite{yang2020learning, guo2019subspace}.

\textbf{Metrics.} In all empirical studies, we conduct untargeted attack until iteration $N=10000$, i.e., the maximum query times, and report the victim's remaining accuracy on all samples (Acc.), the average query times (A.Q.), and the median query times (M.Q.) for only successful AEs, which is also the setting in \cite{papernot2017practical, guo2019simple, andriushchenko2020square}. A.Q., rather than the total queries, is the main concern for query-based attacks. The reason herein is that A.Q. directly signifies the concealment of attacks. High A.Q. means many queries with similar AEs from the same original sample, which are highly detectable by defenders. By contrast, low A.Q. means queries with samples of mostly different objects, which behave like normal usage of DNNs. Thus, it is reasonable for \cite{ma2020metasimulator, yang2020learning, zhang2020dual} and us to reduce A.Q. by using different query samples to train surrogates. And the data for NAS training is sufficient, e.g., 2 A.Q. for 10K test images in MNIST yields 20K training data for surrogates.

\textbf{Victims.} We attack various victims including WideResNet10 (WRN10 \cite{ZagoruykoK16}), ResNet with Pre-activation (RNP \cite{he2016identity}), and DenseNet40 (DN40 \cite{huang2017densely}) for MNIST, WideResNet28 (WRN28 \cite{ZagoruykoK16}), GDAS \cite{dong2019searching}, and PyramidNet272 (PyrN272 \cite{han2017deep}) for CIFAR10, InceptionV3 (IncV3 \cite{szegedy2016rethinking}), ResNetX101 (RNX101 \cite{xie2017aggregated}), and MNASNet (MNASN \cite{tan2019mnasnet}) for ImageNet. Victims are pre-trained models from Torchvision \cite{paszke2019pytorch} for ImageNet, official repositories for CIFAR10, and our training for MNIST from \cite{pytorchclassification}. To evaluate QueryNet on real-world black-box settings, we also attack 2 commercial APIs from Baidu EasyDL \cite{easydl}, where we have no knowledge about the victims but could only get its output probability. We use PyTorch \cite{paszke2019pytorch} platform with 4 NVIDIA Tesla V100 GPUs, and the report of QueryNet's time consumption is based on these devices.

\begin{table}
\caption{Dataset and the attack bound $\varepsilon$}
\label{setup}
\centering
\renewcommand\tabcolsep{5.5pt}
\begin{tabular}{c|ccccc}
\toprule
Dataset & Image Size & Data & Test Data & $\varepsilon ~(l_\infty)$ & $\varepsilon ~(\ell_2)$ \\ \midrule
MNIST  & $28\times28\times1$ & 70K&10K & 0.3 & 3 \\
CIFAR10 & $32\times32\times3$ & 60K&10K & 16 / 255 & 3  \\
ImageNet & $224\times224\times3$ & 1300K&1K & 0.05 & 5  \\
\bottomrule
\end{tabular}
\end{table}

\textbf{Dataset.} Information about the dataset and the attack bound is specified in Table \ref{setup}, where columns 3 and 4 show the sample number in the dataset and our evaluation, respectively. For ImageNet, we randomly select one image from each of the 1K categories in the validation set and resize the samples into $224$ resolution for the victim's and surrogates' inputs. We use all 10K samples in the MNIST/CIFAR10 test set. To the best of our knowledge, no such large-scale experiment has been conducted for query-based attacks. For attack bounds, we mostly adopt common values (columns 5 and 6 in Table \ref{setup}).

\textbf{Bit-depth.} Existing methods, especially those requiring accurate gradient estimation, generally feed models with 32-bit images \cite{guo2019subspace, yang2020learning, andriushchenko2020square, ma2020metasimulator}. However, real-world models generally accept only 8-bit image inputs. This makes a great difference in performance evaluation, seeing our reproduction of \cite{yang2020learning} using 8-bit and 32-bit queries. According to Table \ref{8-bit}, restricting to 8-bit query input dramatically decreases the performance by a large margin. Thus, it is meaningful and significant for us to conduct large-scale evaluations in this real-world setting. We implement 8-bit setting by cutting the 32-bit query samples to 8-bit in the $[0,255]$ image space, since all samples are rescaled to $[0,1]$ and normalized according to the mean and the variance of the dataset in pre-processing.

\begin{table}
\caption{Influence of the 8-bit setting: study on ImageNet by ResNet50 (clean accuracy 87.91\%)}
\label{8-bit}
\centering
\renewcommand\tabcolsep{5.5pt}
\begin{tabular}{cl|ccc|ccc}
\toprule
      &  & \textbf{Acc.} & \textbf{A.Q.} & \textbf{M.Q.} & \textbf{Acc.} & \textbf{A.Q.} & \textbf{M.Q.}\\
\midrule
    \textbf{$\ell_{\infty}$} & \textbf{RN50} &  \multicolumn{3}{c}{\textbf{32-bit query}} & \multicolumn{3}{c}{\textbf{8-bit query}} \\
\midrule
    \multirow{4}*{$\frac{16}{255}$}
    & SimBA    & 10.2\% & 179.3 & 56 & 29.0\% & 486.7 & 81 \\
    & SimBA+   & 10.1\% & 178.4 & 54 & 26.7\% & 382.5 & 65 \\
    & SimBA++  & 4.0\% & 140.6 & 40 & 29.0\% & 405.3 & 56 \\
    & LeBA     & 1.1\% & 120.6 & 27 & 11.9\% & 538.8 & 45 \\
\midrule
    \multirow{4}*{$\frac{32}{255}$}
     & SimBA   & 9.5\% & 212.7 & 32 & 11.9\% & 208.0 & 36 \\
     & SimBA+  & 6.7\% & 117.6 & 25 & 9.6\% & 158.3 & 34 \\
     & SimBA++ & 0.6\% & 42.3 & 5 & 5.1\% & 124.2 & 23\\
     & LeBA    & 0.1\% & 33.2 & 3 & 0.6\% & 84.2 & 8\\
\bottomrule
\end{tabular}
\end{table}

\textbf{NAS.} In PC-DARTS, we inherit most hyper-parameters in \cite{xu2020pc}, but change the number of layers for the diversity of surrogates. Concretely, we choose \# layers as 6, 8, 10 for three surrogates on CIFAR10, and 3, 4, 5 for smaller surrogates on MNIST. We train by all past query pairs as randomly sampled mini-batches until convergence, i.e., the training MSE loss is lower than 2 after 30 iterations. If the loss does not decrease quickly enough, we also stop the training after 100 batches for time-efficiency, and we alter 100 to 1500 in the first iteration when surrogates should be thoroughly trained from scratch. The batch size is 300 for MNIST and 128 for CIFAR10.

\begin{table*}
\caption{The attack performance on MNIST, CIFAR10, and ImageNet}
\label{cifar10results}
\centering
\renewcommand\tabcolsep{8pt}
\begin{tabular}{cc|ccc|ccc|ccc}
\toprule
      &  & \textbf{Acc.} & \textbf{A.Q.} & \textbf{M.Q.} & \textbf{Acc.} & \textbf{A.Q.} & \textbf{M.Q.} & \textbf{Acc.} & \textbf{A.Q.} & \textbf{M.Q.} \\
\midrule
     & \textbf{MNIST} &  \multicolumn{3}{c}{\textbf{WRN10 (99.41\%)}} & \multicolumn{3}{c}{\textbf{RNP (99.59\%)}} & \multicolumn{3}{c}{\textbf{DN40 (99.46\%)}} \\
\midrule
    \multirow{4}*{$\ell_{\infty}$}
    & Bandits & 9.7\% & 149.7 & 26 & 13.9\% & 56.8 & 22 & 17.7\% & 62.9 & 16 \\
     & PPBA & 7.7\% & 49.6 & 2 & 3.0\% & 205.8 & 2 & 8.4\% & 25.4 & 2 \\
     & Square & 5.1\% & 50.2 & 28 & 1.1\% & 13.9 & 1 & 0.8\% & 46.2 & 22 \\
    & QueryNet & \textbf{0.1\%} & \textbf{3.9} & \textbf{2} & \textbf{0.7\%} & \textbf{7.6} & \textbf{1} & \textbf{0.0\%} & \textbf{5.0} & \textbf{4} \\
\midrule
    \multirow{4}*{$\ell_2$}
    & Bandits & 29.7\% & 963.3 & 28 & 40.2\% & 1894.7 & 39 & 26.4\% & 175.8 & 18 \\
    & PPBA & 7.0\% & 41.4 & \textbf{2} & 9.3\% & \textbf{14.6} & \textbf{2} & 8.9\% & \textbf{4.2} & \textbf{2} \\
    & Square & 3.8\% & 192.0 & 58 & 3.2\% & 411.7 & 90 & 0.5\% & 184.3 & 53 \\
    & QueryNet & \textbf{0.5\%} & \textbf{5.3} & 4 & \textbf{0.4\%} & 18.6 & 5 & \textbf{0.2\%} & 5.9 & 4 \\
\midrule
     & \textbf{CIFAR10} &  \multicolumn{3}{c}{\textbf{WRN28 (95.97\%)}} & \multicolumn{3}{c}{\textbf{GDAS (97.19\%)}} & \multicolumn{3}{c}{\textbf{PyrN272 (98.44\%)}} \\
\midrule
    \multirow{7}*{$\ell_{\infty}$}
    & Bandits & 27.0\% & 785.8 & 66 & 38.2\% & 1177.1 & 127 & 54.0\% & 1166.6 & 145 \\
    & SimBA & 13.3\% & 188.2 & \textbf{1} & 1.2\% & 50.5 & 1 & 32.9\% & 401.3 & 20 \\
    & LeBA* & 5.1\% & 164.2 & 2 & 0.7\% & 24.0 & 1 & 18.8\% & 304.6 & 2 \\
    & PPBA & 1.4\% & 78.6 & 2 & 1.1\% & 98.1 & 2 & 3.6\% & 202.1 & 2 \\
    & Subspace* & 0.0\% & 65.0 & 24 & 0.0\% & 53.0 & 24 & 0.2\% & 136.6 & 34 \\
    & Square & 0.0\% & 53.1 & 13 & 0.0\% & 15.6 & 1 & 0.0\% & 56.4 & 4 \\
    & QueryNet & \textbf{0.0\%} & \textbf{16.0} & 2 & \textbf{0.0\%} & \textbf{5.0} & \textbf{1} & \textbf{0.0\%} & \textbf{28.3} & \textbf{2} \\
\midrule
    \multirow{7}*{$\ell_2$}
    & Bandits & 22.2\% & 329.9 & 20 & 18.5\% & 952.8 & 24 & 22.3\% & 1836.3 & 552 \\
    & SimBA & 1.2\% & 81.4 & 7 & 0.2\% & 12.9 & 1 & 7.0\% & 282.9 & 27 \\
    & LeBA* & 0.8\% & 57.1 & 2 & 0.2\% & 8.5 & \textbf{1} & 6.0\% & 235.9 & 2 \\
    & PPBA & 0.2\% & 36.2 & 2 & 0.5\% & 84.4 & 2 & 0.4\% & 132.2 & \textbf{2} \\
    & Subspace* & 0.1\% & 524.9 & 418 & 0.0\% & 474.7 & 440 & 0.0\% & 583.9 & 540 \\
    & Square & 0.0\% & 70.0 & 22 & 0.0\% & 40.4 & 16 & 0.0\% & 106.2 & 63 \\
    & QueryNet & \textbf{0.0\%} & \textbf{9.5} & \textbf{2} & \textbf{0.0\%} & \textbf{5.8} & 2 & \textbf{0.0\%} & \textbf{24.3} & 3 \\

\midrule
     & \textbf{ImageNet} &  \multicolumn{3}{c}{\textbf{IncV3 (77.45\%)}} & \multicolumn{3}{c}{\textbf{RNX101 (79.31\%)}} & \multicolumn{3}{c}{\textbf{MNASN (73.51\%)}} \\
\midrule
    \multirow{5}*{$\ell_{\infty}$}
    & PPBA & 11.8\% & 627.9 & 42 & 7.8\% & 489.1 & 26 & 2.5\% & 284.9 & 10  \\
    & Subspace* & 0.3\% & 114.1 & 22 & 0.0\% & 127.1 & 26 & 0.0\% & 37.5 & 16 \\
    & Square & 0.2\% & 132.1 & 11 & 0.0\% & 82.3 & 15 & 0.0\% & 26.8 & 2 \\
    & QueryNet & 0.3\% & 123.0 & 12 & 0.0\% & 79.8 & 12 & 0.0\% & 24.0 & 2 \\
    & QueryNet* & \textbf{0.2\%} & \textbf{74.7} & \textbf{3} & \textbf{0.0\%} & \textbf{47.3} & \textbf{2} & \textbf{0.0\%} & \textbf{13.5} & \textbf{2} \\
\midrule
    \multirow{5}*{$\ell_2$}
    & PPBA & 19.8\% & 1333.9 & 540 & 4.6\% & 939.2 & 390 & 2.3\% & 683.9 & 291 \\
    & Subspace* & 3.7\% & 1395.0 & 992 & 1.0\% & 1717.5 & 1289 & 0.2\% & 1250.5 & 971 \\
    & Square & 3.6\% & 528.7 & 120 & 0.4\% & 630.6 & 160 & 0.1\% & 442.8 & 129 \\
    & QueryNet & 2.8\% & 518.6  & 72 & 0.5\% & 481.8 & 96 & \textbf{0.0\%} & 280.1 & 50 \\
    & QueryNet* & \textbf{1.9\%} & \textbf{263.5} & \textbf{4} & \textbf{0.2\%} & \textbf{196.0} & \textbf{4} & 0.1\% & \textbf{125.1} & \textbf{3} \\

\midrule
     & \textbf{EasyDL} &  \multicolumn{3}{c}{\textbf{MNIST (98.35\%)}} & \multicolumn{3}{c}{\textbf{CIFAR10 (96.18\%)}} & \multicolumn{3}{c}{\textbf{}} \\
\midrule
    \multirow{2}*{$\ell_{\infty}$}
    & Square & 0.0\% & 3.4 & 1  & 0.0\% & 23.7 & 5 &  \\
    & QueryNet & \textbf{0.0\%} & \textbf{2.4} & \textbf{1}  & \textbf{0.0\%} & \textbf{16.0} & \textbf{2} & & &\\
\midrule
    \multirow{2}*{$\ell_2$}
    & Square & 0.0\% & 31.6 & 19 & 0.0\% & 28.1 & 9 &  \\
    & QueryNet & \textbf{0.0\%} & \textbf{21.5} & \textbf{7}  & \textbf{0.0\%} & \textbf{14.4} & \textbf{2} & & &\\
\bottomrule
\end{tabular}
\end{table*}

\begin{table*}
\caption{Ablation study on different attack bounds (CIFAR10, WRN28)}
\label{Ablationstudy}
\centering
\renewcommand\tabcolsep{8pt}
\begin{tabular}{cc|ccc|ccc|ccc|ccc}
\toprule
     & &\multicolumn{3}{c}{\textbf{Square Baseline}} & \multicolumn{3}{c}{\textbf{Ours $\backslash$ (NAS \& Square+)}} & \multicolumn{3}{c}{\textbf{Ours $\backslash$ (NAS)}} & \multicolumn{3}{c}{\textbf{Ours (QueryNet)}} \\
\midrule
    & $\epsilon$ & \textbf{Acc.} & \textbf{A.Q.} & \textbf{M.Q.} & \textbf{Acc.} & \textbf{A.Q.} & \textbf{M.Q.} & \textbf{Acc.} & \textbf{A.Q.} & \textbf{M.Q.} & \textbf{Acc.} & \textbf{A.Q.} & \textbf{M.Q.}\\
\midrule
     \multirow{5}*{$\ell_\infty$} & 4 / 255 & 4.3\% & 1372 & 675 & 3.8\% & 1300 & 612 & 3.7\% & 1292 & 595 & \textbf{3.2\%} & \textbf{1103} & \textbf{380} \\
     & 8 / 255 & 0.2\% & 365.6 & 119 & 0.1\% & 336.0 & 89 & 0.1\% & 323.5 & 89 & \textbf{0.1\%} & \textbf{210.0} & \textbf{30} \\
     & 12 / 255 & 0.0\% & 129.4 & 37 & 0.0\% & 102.9 & 20 & 0.0\% & 100.0 & 19 & \textbf{0.0\%} & \textbf{47.6} & \textbf{3} \\
     & 16 / 255 & 0.0\% & 53.1 & 13 & 0.0\% & 35.3 & 3 & 0.0\% & 33.6 & 3 & \textbf{0.0\%} & \textbf{16.0} & \textbf{2} \\
     & 20 / 255 & 0.0\% & 27.1 & 4 & 0.0\% & 15.5 & 2 & 0.0\% & 15.6 & 2 & \textbf{0.0\%} & \textbf{6.7} & \textbf{2} \\
\midrule
     \multirow{5}*{$\ell_2$} & 1 & 3.1\% & 807.5 & 256 & 3.1\% & 778.6 & 240 & 3.3\% & 762.0 & 245 & \textbf{2.4\%} & \textbf{554.2} & \textbf{96} \\
     & 2 & 0.2\% & 194.6 & 53 & 0.1\% & 168.2 & 41 & 0.1\% & 155.8 & 38 & \textbf{0.0\%} & \textbf{26.8} & \textbf{4}\\
     & 3 & 0.0\% & 70.0 & 22 & 0.0\% & 44.0 & 6 & 0.0\% & 42.7 & 7 & \textbf{0.0\%} & \textbf{9.5} & \textbf{2} \\
     & 4 & 0.0\% & 34.0 & 10 & 0.0\% & 16.7 & 3 & 0.0\% & 14.0 & 3 & \textbf{0.0\%} & \textbf{5.0} & \textbf{2} \\
     & 5 & 0.0\% & 17.5 & 4 & 0.0\% & 8.5 & 2 & 0.0\% & 7.0 & 2 & \textbf{0.0\%} & \textbf{3.5} & \textbf{2} \\
\bottomrule
\end{tabular}
\end{table*}

\begin{figure*}
    \centering
    \includegraphics[width=\hsize]{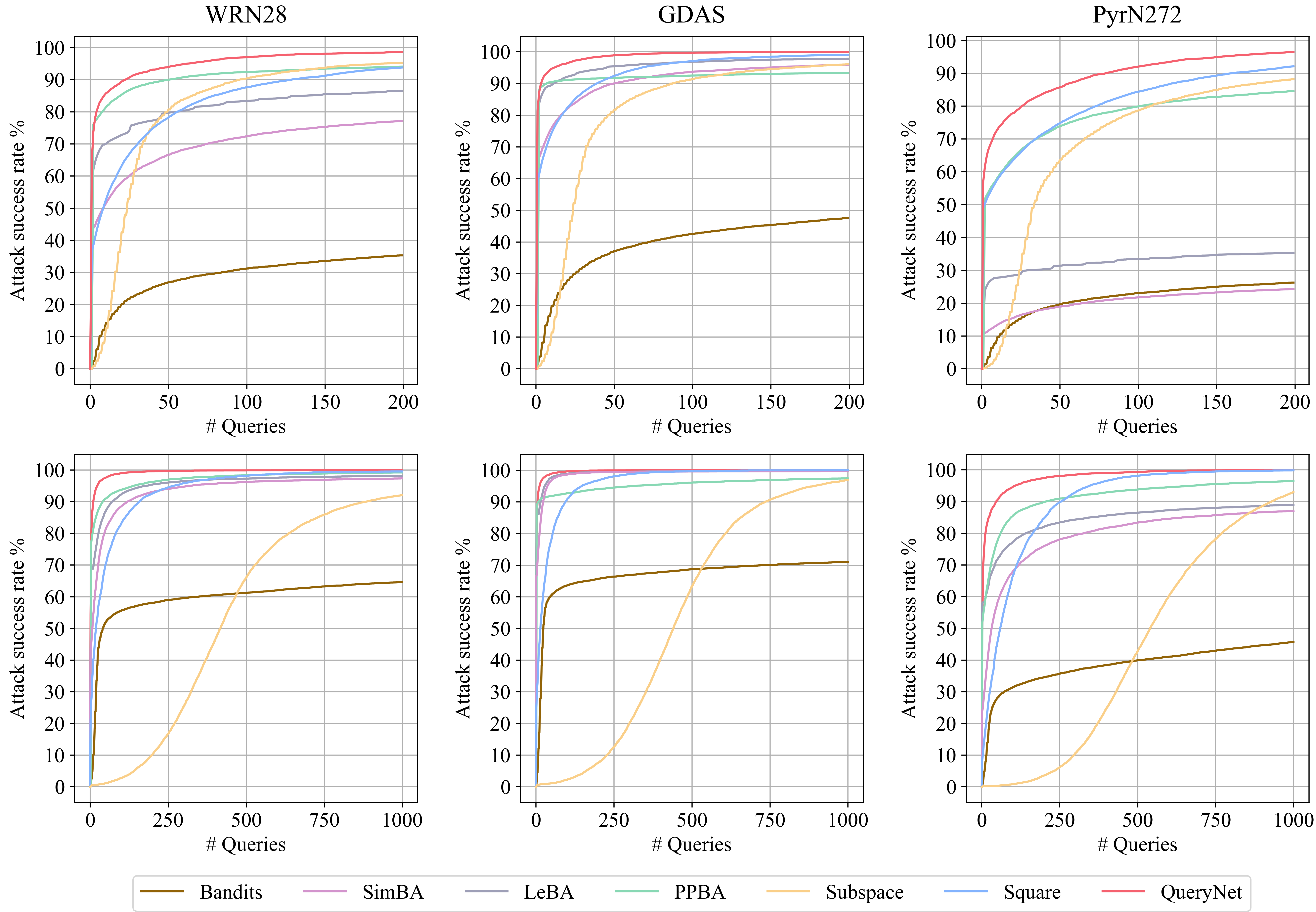}
    \caption{The trend of attack success rates for different methods on CIFAR10 in Table \ref{cifar10results}. Top figures are for $\ell_{\infty}$ attacks and bottom ones are for $\ell_2$ attacks.}
    \label{fig:ResultsonCIFAR10}
\end{figure*}

\begin{figure*}
    \centering
    \includegraphics[width=0.48\hsize]{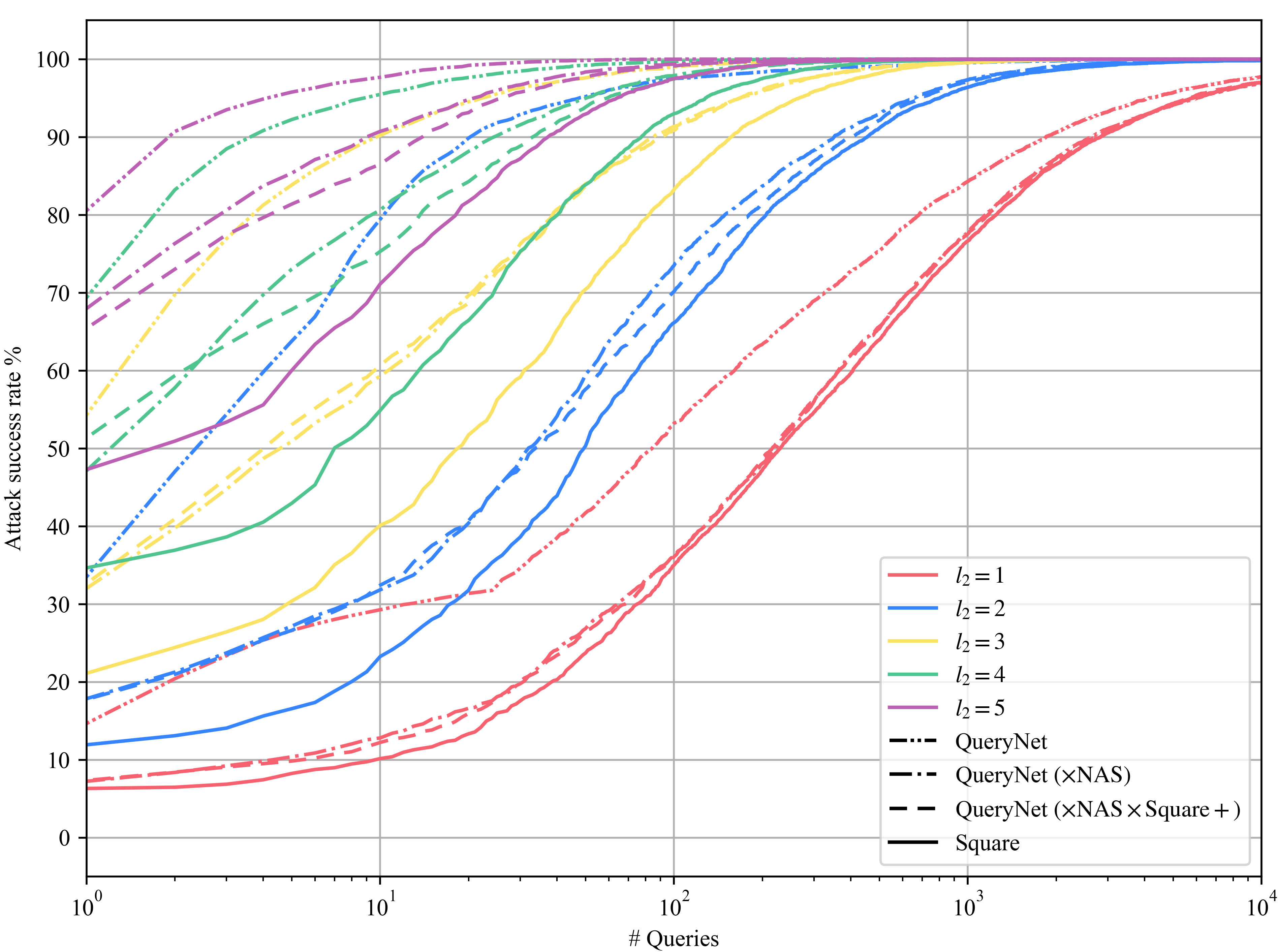}
    \includegraphics[width=0.48\hsize]{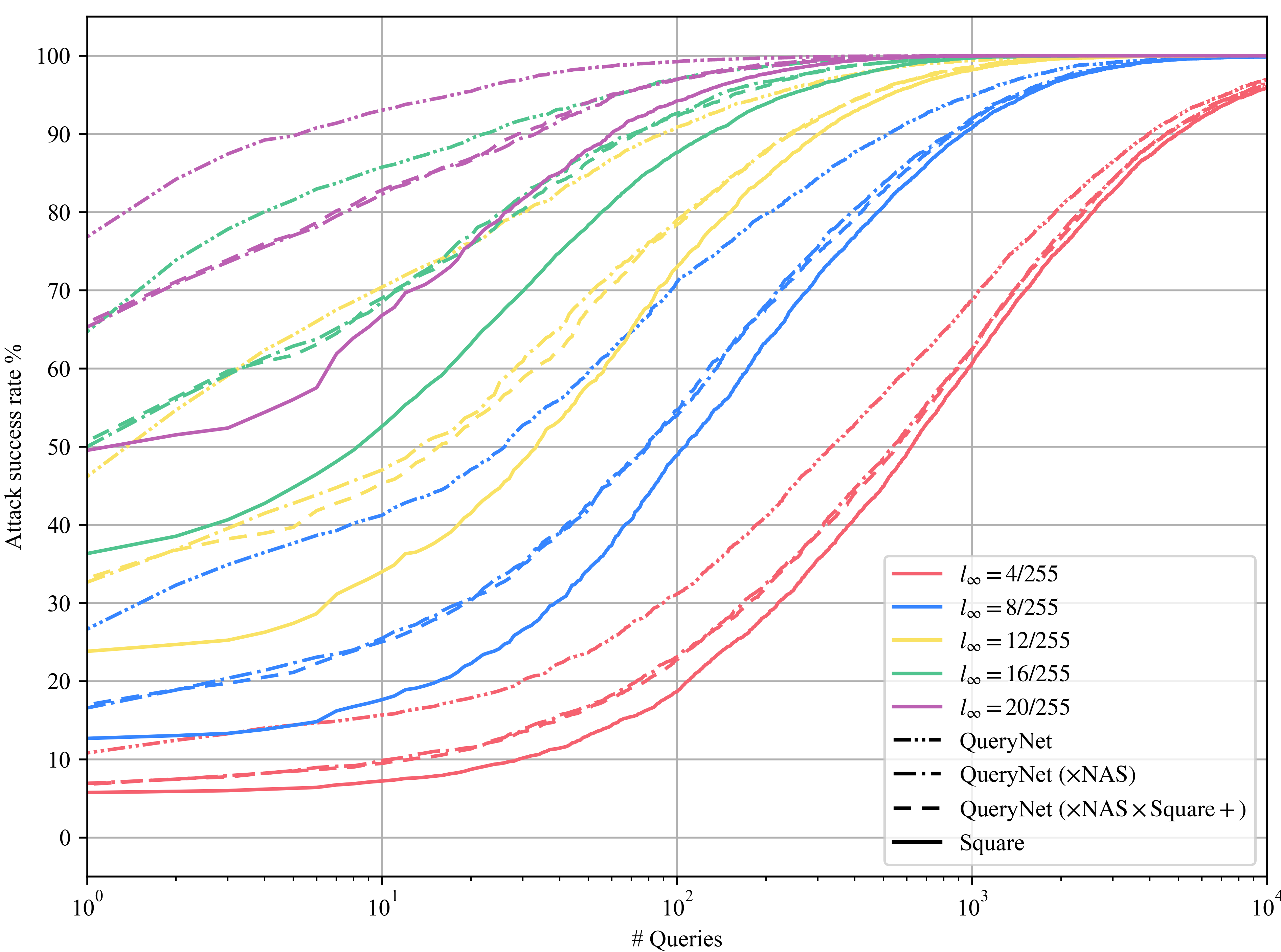}
    \caption{The trend of attack success rates for Table \ref{Ablationstudy}, in logarithmic axis.}
    \label{fig:Resultsofablationstudy}
\end{figure*}

\subsection{Results}
The attack performance is reported in Table \ref{cifar10results}, where ``*'' indicates the method uses surrogates pre-trained on the victim's training set. Overall, QueryNet achieves a significant reduction in queries from the SOTA attacks, while maintaining the highest success rates. On MNIST, the A.Q. decreases from 192.0 to an amazing 5.3 in attacking WRN10, surpassing baselines by an order of magnitude. QueryNet reduces the victim's accuracy from over 99.4\% to below 0.7\% in all conducted evaluations. And QueryNet also greatly surpasses BayesOpt \cite{shukla2019black} based on our assessment. Note that the attack performance is not good when A.Q. is small with a high Acc., e.g., PPBA in MNIST, because A.Q. and M.Q neglects failed AEs.

On CIFAR10, a similar trend can be observed. QueryNet cuts down about 50\% to 70\% queries in $\ell_\infty$ attacks and 75\% to 85\% queries in $\ell_2$ attacks from Square attack, which is already verified excellent. In all evaluations, QueryNet achieves a definite fooling within 30 A.Q. for a sample. The reduction of queries by QueryNet is significant in $\ell_2$ attacks, indicating that DNNs are vulnerable to $\ell_\infty$-unrestricted AEs. We also visualize our searched architecture in Appendix \ref{apd:arch}.

QueryNet's impact on decreasing A.Q. on ImageNet is also obvious, especially in $\ell_2$ attacks. It also decreases the M.Q., reducing the victim's accuracy very fast in initial iterations. Due to the computational consumption of NAS on ImageNet, we train the surrogates only once and do not finetune them later, thus only revealing QueryNet's partial potential here. Still, QueryNet outperforms all baselines, especially in $\ell_2$ attacks. For further reference, we also report results with pre-trained surrogates (DN121 \cite{huang2017densely}, DN169 \cite{huang2017densely}, and RN50 \cite{he2016deep}) as QueryNet*, but that is not our main focus.

Baidu EasyDL \cite{easydl} provides a platform for black-box attack evaluation, which is fair and more convincing, since no one can access the neural networks' knowledge. For those models in Baidu EasyDL, both Square and QueryNet achieve definite fooling quickly, but the required number of queries in QueryNet is significantly smaller. Our experiments demonstrate that QueryNet remains very query-efficient in complete black-box settings, where we have no idea how the victims are built and trained. This experiment sounds alarmed that the existing commercial models on the platform are not robust and can be attacked within only a few queries, so that the defender \cite{li2020blacklight, qin2021random, chen2022adversarial, wu2022unifying} is hard to detect attacks based on the query times.

For a deeper illustration, we present the attack success rates after every query iteration in Fig. \ref{fig:ResultsonCIFAR10}. In all performed experiments, QueryNet achieves high success rates very quickly in initial queries, resulting in low M.Q., and also efficiently modifies the remaining robust samples for the victim, leading to low A.Q. and Acc. The striking performance is obtained by our novel design to focus on surrogates' PS and GS.

\subsection{Ablation study}
To investigate the influence of our major designs, we here conduct the ablation studies under different attack bounds. First, we exclude Square+ and NAS from QueryNet and report the performance using trainable surrogates (DenseNet121, DenseNet169, and ResNet50) in column 3 of Table \ref{Ablationstudy}. One could observe that our strategy of surrogates' multiple identities, ensemble, and learnable self-evaluation, and adjustable surrogates, could significantly reduce queries from Square attack baseline. Second, we additionally equip the Square+ attacker, and the attack performance is further improved (column 4). Finally, the complete QueryNet (column 5) obtains dramatic improvement, showing that NAS could lead to much better GS/PS and thus greatly contribute to the quality of surrogates. The success rate is plotted in Fig. \ref{fig:Resultsofablationstudy}. Every four lines with the same color also show that our novel designs all positively contribute to the final amazing performance.

To investigate the influence of the number of surrogates $n$, we construct different numbers of diverse surrogates by altering the number of layers in PC-DARTS, and perform each evaluation 5 times to reduce interference from randomness. The results are reported in Table \ref{abls}. A natural conclusion here is that more surrogates lead to better and more stable attacks, but require more computation. For a trade-off, we select $n=3$ in all remaining experiments in this paper. Notice that in the previous experiments, only one trial is tested.
\begin{table}
\caption{The attack performance and GPU hours with different $n$ values (CIFAR10, WRN28, $\ell_\infty=\frac{16}{255}, \ell_2=3$)}
\label{abls}
\centering
\begin{tabular}{c|cc|cccc}
\toprule
    & $n$ & \# Layers & \textbf{Acc.} & \textbf{A.Q.} & \textbf{M.Q.} & \textbf{GPU}\\
\midrule
     \multirow{3}*{$\ell_{\infty}$} & 1 & 10 & 0.0\% & $16.85\pm0.59$ & 2 & 0.8 \\
     & 3  & 6,8,10& 0.0\% & $14.61\pm0.29$ & 2 & 2.3 \\
     & 5 & 4,6,...,12 & 0.0\% & $13.29\pm0.13$ & 2 & 8.9 \\
\midrule
     \multirow{3}*{$\ell_2$} & 1 & 10 & 0.0\% & $12.65\pm0.29$ & 3 & 1.1\\
     & 3  & 6,8,10 & 0.0\% & $8.76\pm0.25$ & 2 & 3.7\\
     & 5 & 4,6,...,12 & 0.0\% & $7.56\pm0.22$ & 2 & 15.7\\
\bottomrule
\end{tabular}
\end{table}

\subsection{Computation analysis}\label{simulator}
For adversarial attacks, the computation time is not very strict \cite{athalye2018obfuscated}. The proposed QueryNet indeed needs computation to perform NAS. But QueryNet can be well implemented using parallel computing on data and model with early-stopping, then the overall time consumption is still acceptable. As in Table \ref{computation}, QueryNet crafts one AE under common bounds within only 0.5 to 2.6 GPU seconds, i.e., 0.15s to 0.65s using 4 GPUs on average. In comparison, the recent Simulator attack \cite{ma2020metasimulator} requires over 72 GPU hours to train the meta attacker before the first query. Also, ZOO \cite{chen2017zoo} consumes an average of 56.35 seconds to successfully attack one CIFAR-10 image on WRN28, while QueryNet only needs 1.40 seconds. And Bandits, Subspace, LeBA, PPBA, and SimBA all conduct no data paralleling and is thus much slower than QueryNet. Therefore, QueryNet has no significant disadvantage on computation time.

\begin{table}
\caption{The time consumption of QueryNet, measured by total GPU hours and GPU seconds per image}
\label{computation}
\centering
\renewcommand\tabcolsep{5pt}
\begin{tabular}{c|c|ccc|cc}
\toprule
   & Victim & \textbf{Acc.} & \textbf{A.Q.} & \textbf{M.Q.} & \textbf{GPU}  & \textbf{per image}\\
\midrule
   \textbf{MNIST} & \textbf{WRN10} & 0.1\% & 3.9 & 2  &  6.1 hrs & 2.20 s\\
   $\ell_\infty=0.3$ & \textbf{RNP} & 0.7\% & 7.6& 1 &  2.4 hrs & 0.86 s\\
    & \textbf{DN40} & 0.0\% & 5.0 & 4 &  1.6 hrs & 0.58 s\\
\midrule
   \textbf{CIFAR10} & \textbf{WRN28} & 0.1\% & 210 & 30  & 3.9 hrs & 1.40 s\\
   $\ell_\infty=\frac{8}{255}$ & \textbf{GDAS} & 0.0\% & 62 & 6 & 4.3 hrs & 1.55 s\\
    & \textbf{PyrN272} & 0.0\% & 219 & 70 & 7.2 hrs & 2.59 s \\
\bottomrule
\end{tabular}
\end{table}

\section{Conclusion and future work}\label{conclusion}
We develop the QueryNet to exploit and improve both the GS and PS of surrogates to achieve great attack performance. The query-efficient comes from the novel usage of multi-identity surrogates as transferable attackers for their GS and transferability evaluators due to the PS. In our unified framework, surrogates are also creatively trained by NAS, further improving the performance. QueryNet reduces queries by about an order of magnitude while maintaining the highest fooling rate in our comprehensive and realistic evaluations.

Note that QueryNet cannot be applied to decision-based attacks by training surrogates with hard labels, because in this case, QueryNet still needs the model's soft outputs to greedily decide whether a query sample is more adversarial. And in the future, it would be interesting to explore the surrogate's usage in decision-based attacks. Also, it is possible to design other attackers on QueryNet, e.g., by adopting more extensible architecture space, more transferable white-box attacks, and more advanced NAS strategies. However, our attack is only applicable in digital space and thus does not exert imminent impacts to the deployed DNNs on life-concerning systems like physical attacks, which demand a stable invasion across cameras and sensors.



\bibliographystyle{IEEEtran}
\bibliography{Reference}

\begin{IEEEbiography}[{\includegraphics[width=1in,height=1.25in,clip,keepaspectratio]{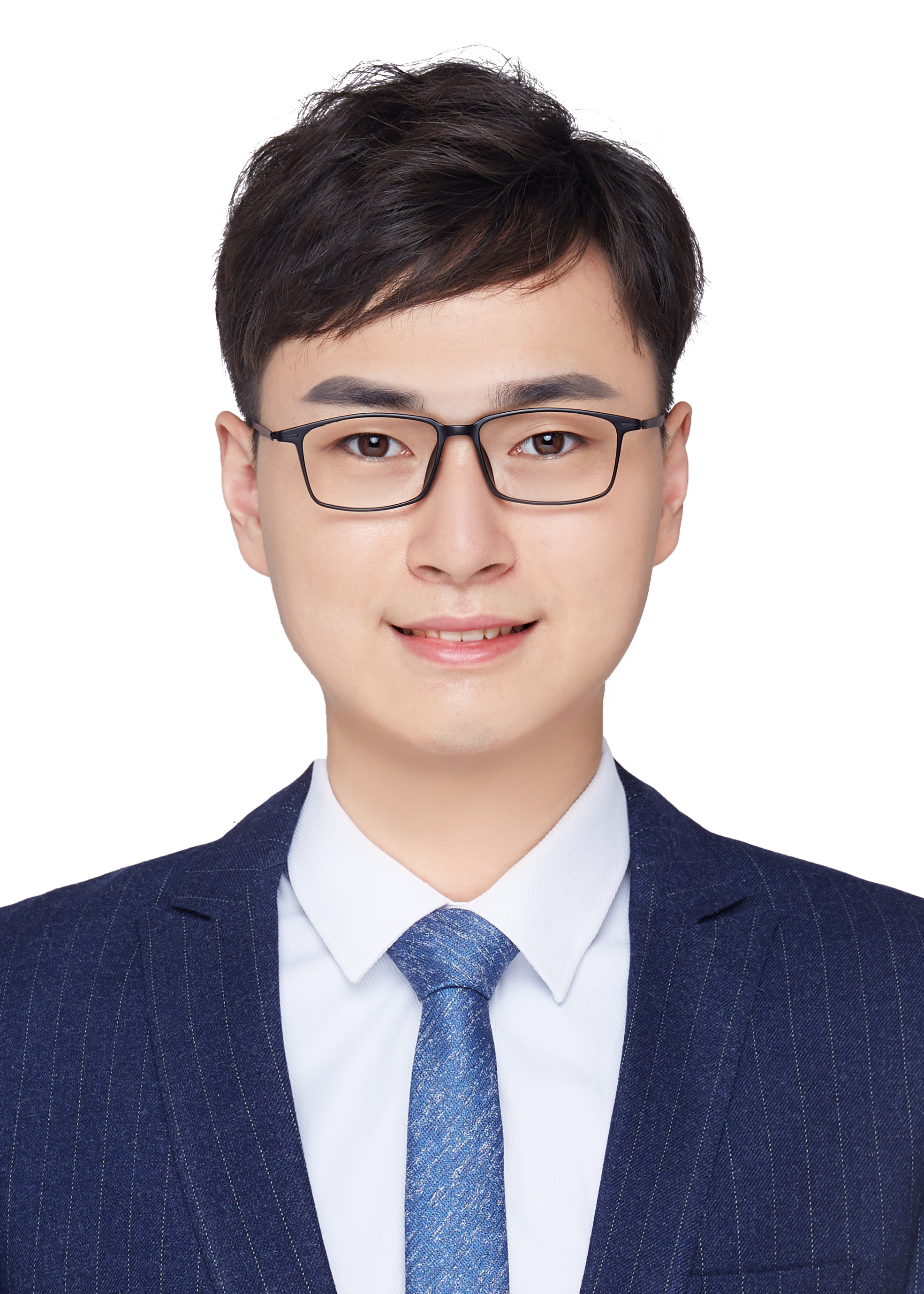}}]{Sizhe Chen}
received his B.E. and M.E. degree in Shanghai Jiao Tong University, Shanghai, China, in 2020 and 2023, respectively. He is a Ph.D. student at the Department of Electrical Engineering and Computer Sciences, University of California, Berkeley. His research interests are focused on trustworthy machine learning.
\end{IEEEbiography}

\begin{IEEEbiography}[{\includegraphics[width=1in,height=1.25in,clip,keepaspectratio]{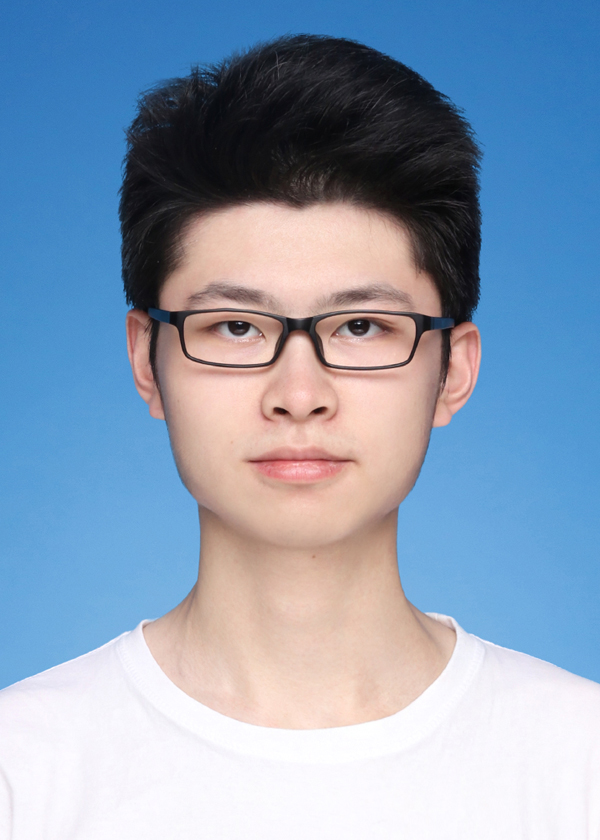}}]{Zhehao Huang}
received his B.E. degree in Shanghai Jiao Tong University, Shanghai, China, in 2022. He is now a Ph.D. student at the Institute of Image Processing and Pattern Recognition, Shanghai Jiao Tong University, Shanghai, China. His research interests are architecture design and sparsity of DNNs.
\end{IEEEbiography}

\begin{IEEEbiography}[{\includegraphics[width=1in,height=1.25in,clip,keepaspectratio]{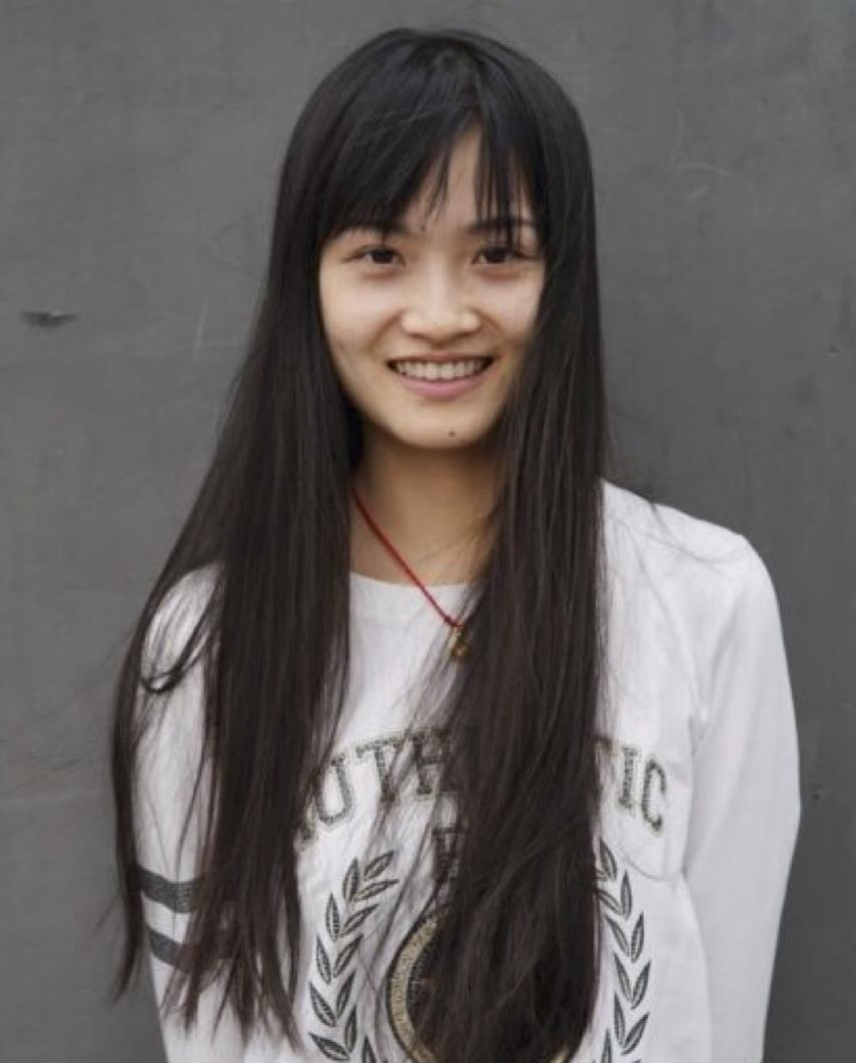}}]{Qinghua Tao}
received the B.S. degree from Central South University, China, in 2014, and the Ph.D. degree from Tsinghua University, China, in 2020. She is currently a Post-Doctoral Researcher with ESAT-STADIUS, KU Leuven, Belgium. Her research interests include machine learning, dynamic systems and optimization, especially for the analysis and applications of piecewise linear neural networks.
\end{IEEEbiography}

\begin{IEEEbiography}[{\includegraphics[width=1in,height=1.25in,clip,keepaspectratio]{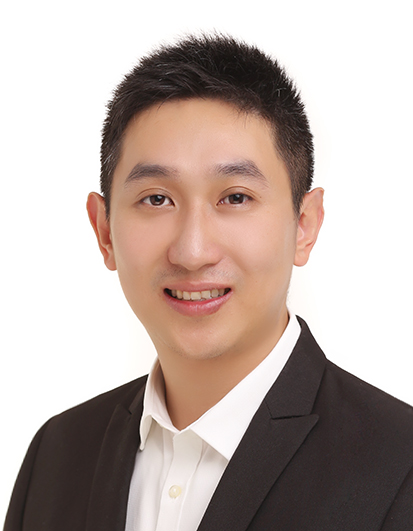}}]{Xiaolin Huang}
(S'10-M'12-SM'18) received the B.S. degree in control science and engineering, and the B.S. degree in applied mathematics from Xi'an Jiaotong University, Xi'an, China in 2006. In 2012, he received the Ph.D. degree in control science and engineering from Tsinghua University, Beijing, China. From 2012 to 2015, he worked as a postdoctoral researcher in ESAT-STADIUS, KU Leuven, Leuven, Belgium. After that he was selected as an Alexander von Humboldt Fellow and working in Pattern Recognition Lab, the Friedrich-Alexander-Universit\"{a}t Erlangen-N\"{u}rnberg, Erlangen, Germany. From 2016, he has been an Associate Professor at Institute of Image Processing and Pattern Recognition, Shanghai Jiao Tong University, Shanghai, China. In 2017, he was awarded by "1000-Talent Plan" (Young Program).

His current research areas include machine learning and optimization, especially for robustness and sparsity of both kernel learning and deep neural networks.
\end{IEEEbiography}

\clearpage
\appendix

\subsection{Attackers in QueryNet}\label{attackdetails}
In the forward propagation of QueryNet, one could flexibly choose diverse attackers. Here we specify a feasible and efficient design of attackers, which contain model-dependent and model-independent attacks.

Model-dependent attacks are to attack surrogates and generate AEs based on the attack transferability. Here we want a fast attack to accelerate the fooling, and thus choose FGSM \cite{goodfellow2014explaining}, which is also preferred due to its good transferability \cite{su2018robustness}. Multi-step attacks, like Projected Gradient Descend \cite{madry2017towards} and its transferability enhance techniques \cite{dong2018boosting, dong2019evading}, have been tested in QueryNet but are not as efficient as FGSM. With FGSM, line 1 in Alg. \ref{amida} is specified as
\begin{eqnarray*}\label{fgsm}
\begin{split}
X_i =
\left\{
\begin{array}{ll}
\text{clip}_{X^\mathrm{org}}^\varepsilon(X + 2 \varepsilon \cdot \text{sign} (\nabla_{\boldsymbol x} \mathcal{L}(\mathcal{S}_i(X), Y))) ,  &p = \infty, \\ 
\text{rescale}_{X^\mathrm{org}}^\varepsilon(X + 2 \varepsilon \cdot  \frac{\nabla_{\boldsymbol x} \mathcal{L}(\mathcal{S}_i(X), Y)}{\|\nabla_{\boldsymbol x} \mathcal{L}(\mathcal{S}_i(X), Y)\|_2}) ,   &p = 2, \\
\end{array}
\right.
\end{split}
\end{eqnarray*}
where $\text{clip}(\cdot)$ denotes that the pixels being directly clipped into the bound $\varepsilon$ as in \cite{madry2017towards}, and $\text{rescale}(\cdot)$ means the perturbation $X'-X$ is rescaled to the $\ell_2$ bound as in \cite{andriushchenko2020square, yang2020learning}. We modify the step size to $2\varepsilon$ to ensure that every pixel reaches the bound at all iterations, enhancing the transferability \cite{chen2021going}.

Model-independent attacks, mainly referring to random-search attacks \cite{guo2019simple, andriushchenko2020square, li2020projection}, do not rely on surrogates. In our study, we select the excellent SOTA Square attack \cite{andriushchenko2020square}, which adds a random-generated square to the current perturbation in each iteration. Square attack initializes the perturbation by random vertical stripes to boost the fooling, which is also adopted in QueryNet.

In this random-search attack, past query pairs from the same original sample are not taken into account for reducing queries. To take full use of this information, we resort to the sequential search approach for black-box optimization problems, where the goal, similar to the iterative query, is to find the optimum with a small budget of function evaluation times. To achieve that, \cite{malherbe2017global} develops a mechanism to evaluate only the promising samplings, termed potential maximizers. Inspired by that, we propose Square+ to yield a candidate AE only if it is the potential maximizer. Otherwise, we retry the random search on the square location until it reaches a maximum times $M$. A sample $x^q$ (labeled $y$) is a potential maximizer, according to \cite{malherbe2017global}, if and only if it satisfies the Lipschitz continuity assumption on the victim's loss, i.e.,
\begin{eqnarray}\label{maximizer}
\begin{split}
& \max\limits _{i \neq j} \frac{\left|\mathcal{L}(\mathcal{V}(\boldsymbol q_{i}),y)-\mathcal{L}(\mathcal{V}(\boldsymbol q_{j}),y)\right|}{\left\|\boldsymbol q_{i}-\boldsymbol q_{j}\right\|_{2}}
\left\|\boldsymbol x^q-\boldsymbol q_{i}\right\|_{2} - \\
& ~~~~~~~~~~~ \max\limits _{i} \mathcal{L}(\mathcal{V}(\boldsymbol q_{i}),y)
\geq - \beta \min\limits _{i} \mathcal{L}(\mathcal{V}(\boldsymbol q_{i}),y),
\end{split}
\end{eqnarray}
where $\boldsymbol q_i$ is the $i^\text{th}$ queried AE for the original sample of $\boldsymbol x^q$, and $\beta$ is the hyper-parameter to decide the tightness of this constraint. In our implementation, we heuristically choose $\beta = 0.7$ and $M = 50$. The first term in (\ref{maximizer}) involves the estimated Lipschitz constant from all past query samples. Aware on that, Square+ improves the attack diversity and boosts the query-efficiency further, seeing our ablation study.

\subsection{The attacker evaluation weights}\label{midaalg}
In the backward propagation of QueryNet, the surrogates' weights for ensemble and evaluation $\boldsymbol w$ can also be calibrated by the query information. The weight reflects the quality of a surrogate. Accordingly, we set it as the ratio of its yielded AEs that become more adversarial to $V$ as
\begin{eqnarray}\label{weights}
w_i = \frac{\#_{\boldsymbol a_I} (i)}{\#_{\boldsymbol a}(i)}, ~ I = \{k: \mathcal{L}(\mathcal{V}(X^q_k),Y_k^\mathrm{org}) < L_k \},
\end{eqnarray}
where $\boldsymbol a$ is the selected attacker index in the forward propagation (\ref{update}), and $\#_{\boldsymbol a}(i)$ denotes the times that $i$ appears in $a$, i.e., the number of $X^q$ that come from candidate AEs $X_i$ crafted by $A_i$. $\#_{\boldsymbol a_I} (i)$ refers to the number of $X^q$ from $X_i$ that become more adversarial to the victim, because the set $I$ contains the indexes of the query samples that decrease the victim's loss $L$. Generally, if AEs from one attacker are frequently selected, its evaluation weight would be large. In an extreme case, $w_i=0$ if $\#_{\boldsymbol a}(i)=0$, i.e., no candidate AE from $S_i$ is selected, so its quality is low. The evaluation weight for a surrogate signifies its quality compared to others. Thus, in the beginning, all $n$ surrogates are considered equally important with the same weight, namely $[w_1, w_2,...w_n] = \boldsymbol 1$.

\begin{figure}
    \centering
    \includegraphics[width=\hsize]{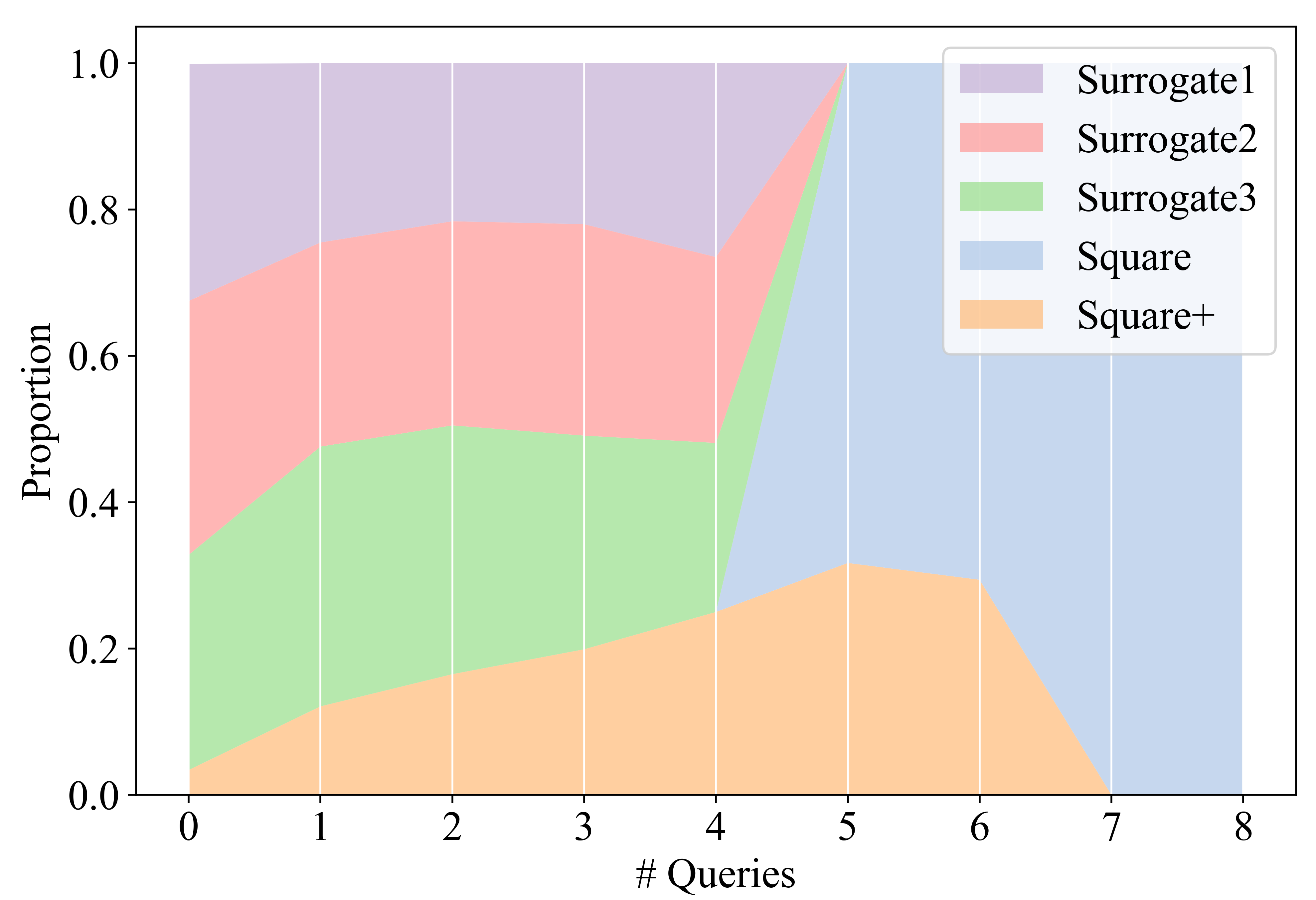}
    \caption{Proportion of functional attackers. In iteration $0 \to 4$, we adopt surrogate attackers and Square+. As surrogates' transferability effect decreases, QueryNet automatically switches to Square+ and Square in iteration $4 \to 7$. Finally, the Square takes the lead, generating the most effective AEs, and are thus used solely after iteration 7. The results are produced by WRN28 on CIFAR10 with $\ell_2=3$.}
    \label{fig:weights}
\end{figure}

By calibrating surrogates' quality, QueryNet's AE evaluation mechanism (\ref{update}) becomes more accurate. Besides evaluation, it is interesting to notice that (\ref{weights}) is also applicable to measure the quality of model-independent attackers. Based on this, we design a strategy to balance different attackers for good query-efficiency and time-efficiency. We initialize the attacker weight for Square+ (attacker $n+1$) and Square (attacker $n+2$) as zero. In the beginning, the prior knowledge is important, so we adopt surrogate attackers and Square+ since they are based on past query information. After the surrogate attackers' quality becomes lower than random-search attacks, we resort to Square+ and Square. Finally, the estimated Lipschitz in Square+ would be also inaccurate \cite{malherbe2017global}, and we use only Square, where it would be unnecessary to evaluate candidate AEs generated from one method, so the computational costly training of surrogates is terminated. This strategy could be expressed as $\{\mathcal{A}_i\}=$
\begin{eqnarray*}\label{balance}
\begin{split}
\left\{
\begin{array}{ll}
\mathcal{A}_1 \to \mathcal{A}_{n+1} ,  & w_{n+1} < \max (w_1 \to w_n), \\
\mathcal{A}_{n+1}, \mathcal{A}_{n+2} ,   & \max (w_1 \to w_n) < w_{n+1}, w_{n+1} < w_{n+2},\\
\mathcal{A}_{n+2},   & w_{n+2} > \max (w_1 \to w_{n+1}).
\end{array}
\right.
\end{split}
\end{eqnarray*}
We illustrate the process above by an example in Fig. \ref{fig:weights}. QueryNet adaptively selects the most efficient attacker for every AE, so different methods work on a dynamic proportion of samples. Although this strategy is heuristically designed, it generalizes to different settings without careful modification.


\subsection{Hyper-parameters for baselines}\label{hyper}
For a fair and unbiased comparison, we inherit most hyper-parameters but carefully tune crucial ones in all compared methods for good performance.

\textbf{Bandits} is a gradient estimation query attack, so we mainly tune the finite-difference step size. Considering its low query-efficiency, we only evaluate the step size 0.1 (original) and 1.0 for CIFAR10 in our 8-bit query setting. The attack success rate is higher at 0.1, while the A.Q. is lower at 1.0. For a trade-off, we set the finite-difference step size 0.1 on both CIFAR10 and MNIST for $\ell_2$ and $\ell_{\infty}$ attacks.

\textbf{SimBA} is a random-search attack. For MNIST, its attack step size 1.0 makes a significant improvement of high attack success rate and low average query times compared with the original step size 0.1, so we choose 1.0 in all evaluations.

\textbf{LeBA} is based on SimBA with additional surrogate models. We set the step size following SimBA and pay more attention to the choice of surrogate models. We comprehensively evaluate ResNet152, DenseNet121, and DenseNet201 for CIFAR10 and select the best-performed ResNet152.

\textbf{PPBA} is a random-search attack. We mainly tune the dimension of the measurement vectors and the change of vector amplitude value $\rho$. We set the dimension as 1500 for ImageNet following the original settings and also for CIFAR10, but we reduce the dimension to 350 for MNIST. We also try to explore the most effective $\rho$ in our 8-bit query setting by evaluating $\rho=0.01,0.03,0.05$ for $\ell_2$ attacks and $\rho=0.01,0.03,0.1,0.3,1.0,3.0$ for $\ell_{\infty}$ attacks. Based on the results, we set the best $\rho=0.01$ for $\ell_2$ attacks and $\rho=1.0$ for $\ell_{\infty}$ attacks in MNIST, CIFAR10, and ImageNet.

\textbf{Subspace} is a gradient estimation query attack with surrogates. We use all pre-trained surrogates in the official implementation, i.e., AlexNet and VGGNets for CIFAR10 and ResNet-18/34/50 for ImageNet. We also evaluate the finite-difference step size of 0.1 (original), 1.0, and 2.0 for CIFAR10 and 0.1, 0.5, 1.0, 3.0 for ImageNet. We finally set step size as 1.0 for CIFAR10 and step size as 3.0 and 1.0 for ImageNet on $\ell_2$ and $\ell_{\infty}$ attacks respectively.

\textbf{Square} is a random-search attack. We inherit the official strategy to gradually decrease the size of squares but set the probability of changing a coordinate $p=0.05$ for all attacks, which is validated better against the original $p=0.1$ in $\ell_2$ attacks.

\subsection{QueryNet adversarial examples}
\begin{figure}
    \centering
    \includegraphics[width=\hsize]{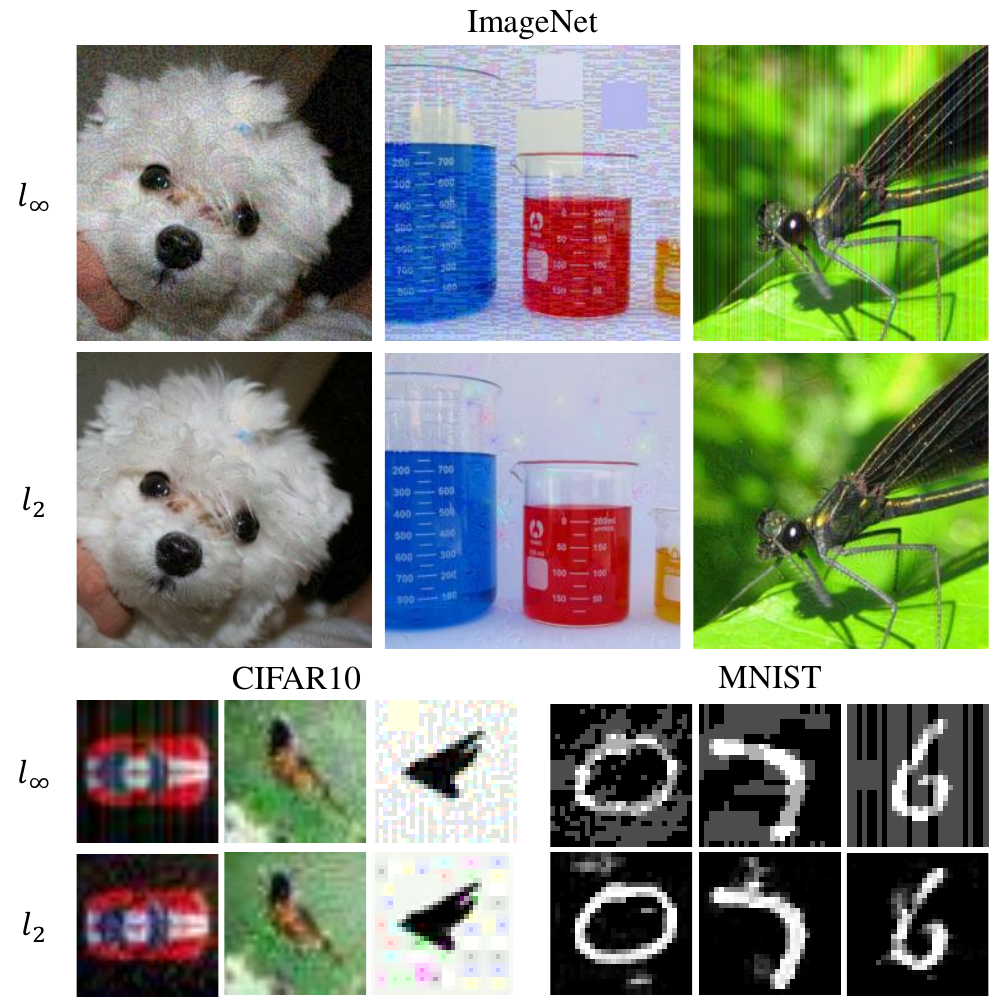}
    \caption{QueryNet AEs. AEs are from attacks on MNASN (ImageNet), EasyDL (CIFAR10), and DN40 (MNIST).}
    \label{fig:adv}
\end{figure}

We visualize the adversarial examples generated by QueryNet in Fig. \ref{fig:adv}, where it is interesting to see the balanced effect of Square attack and FGSM, i.e., the perturbations are constructed by squares and adversarial noise. It illustrates that QueryNet improves the query-efficiency by a dynamic and learnable balance between random-search attack and transfer attack.

\subsection{QueryNet searched architectures}\label{apd:arch}
\begin{figure*}
    \centering
    \includegraphics[width=\hsize]{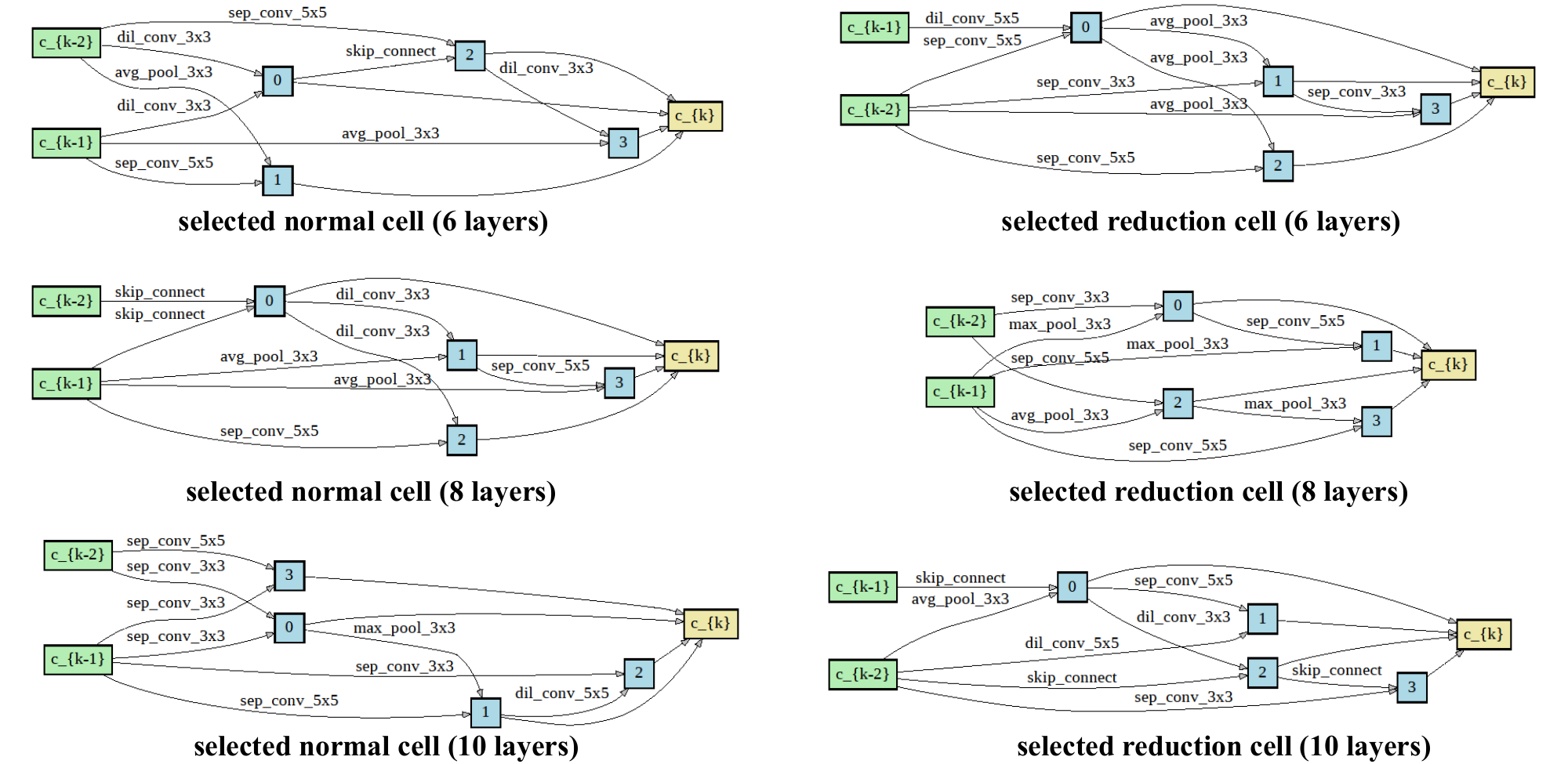}
    \caption{The searched surrogate's architecture by QueryNet when attacking a CIFAR-10 WideResNet28 model to craft 10K samples. All surrogates are trained for 7 attack iterations using randomly sampled data from past query pairs.}
    \label{fig:arch}
\end{figure*}

We visualize surrogates' architectures searched by QueryNet in Fig. \ref{fig:arch}, where we could see that surrogates (with different No. layers) would end with different cells stealing the victim despite training in the same data and searching in the same architecture space. 

QueryNet's success lies in the simultaneous achievement of prediction similarity and gradient similarity via additionally altering the surrogate's architecture. This greater optimization space improves the approximation to the victim, so the query efficiency is significantly boosted despite that we cannot steal the exact architecture merely through observing outputs \cite{cai2023deepguiser}.

\end{document}